\documentclass[journal]{IEEEtran}
\pdfoutput=1

\usepackage{array}
\usepackage[fleqn]{amsmath}
\usepackage{color,soul}
\usepackage{csvsimple}
\usepackage{pgfplots}
\usepackage{filecontents}
\usepackage{amssymb}
\usepackage{amsmath}
\usepackage{url}
\usepackage{mathptmx}
\usepackage{mathtools}
\usepackage{acronym}
\usepackage{booktabs}
\usepackage[compress]{cite}
\usepackage[font={footnotesize}]{caption}
\usepackage[font={footnotesize,up,singlespacing}]{subcaption}
\usepackage[switch,pagewise]{lineno}
\usepackage{graphicx}
\usepackage{tabularx}
\usepackage{times}
\usepackage{mathptmx}
\usepackage[scaled=.85]{helvet}
\usepackage{courier}
\usepackage{color}
\usepackage{tablefootnote}
\usepackage{multicol}
\usepackage{booktabs,caption}
\usepackage[flushleft]{threeparttable}
\usepackage{blindtext}
\usepackage{hyperref}
\hypersetup{
  linkcolor=blue,anchorcolor=black,citecolor=blue,urlcolor=blue
}

\graphicspath{{./} {./img/}}

\graphicspath{{./} {img/}} % Where to find images
\definecolor{fedepink}{cmyk}{0, 0.7808, 0.4429, 0.1412}

\definecolor{ashgrey}{rgb}{0.7, 0.75, 0.71}
\definecolor{babyblueeyes}{rgb}{0.63, 0.79, 0.95}
\sethlcolor{babyblueeyes}

\begin{document}

\title{NullHop: A Flexible Convolutional Neural Network Accelerator Based on Sparse Representations of Feature Maps}

\author{\IEEEauthorblockN{Alessandro Aimar\IEEEauthorrefmark{1},  Hesham Mostafa\IEEEauthorrefmark{1}, Enrico Calabrese\IEEEauthorrefmark{1}, Antonio Rios-Navarro\IEEEauthorrefmark{2}, Ricardo Tapiador-Morales\IEEEauthorrefmark{2}, \\Iulia-Alexandra Lungu\IEEEauthorrefmark{1}, Moritz B. Milde\IEEEauthorrefmark{1}, Federico Corradi\IEEEauthorrefmark{3}, Alejandro Linares-Barranco\IEEEauthorrefmark{2},\\Shih-Chii Liu\IEEEauthorrefmark{1}, Tobi Delbruck\IEEEauthorrefmark{1}}\\
  \IEEEauthorblockA{\IEEEauthorrefmark{1}Institute of Neuroinformatics,
    University of Zurich and ETH Zurich, Switzerland}
    \IEEEauthorblockA{\IEEEauthorrefmark{2}Robotic and Tech. of Computers Lab. University of Seville, Spain}
    \IEEEauthorblockA{\IEEEauthorrefmark{3}iniLabs GmbH, Zurich, Switzerland}}

% make the title area
\maketitle
%TO BE REMOVED BEFORE SUBMIT
\thispagestyle{plain}
\pagestyle{plain}

\begin{abstract}
Convolutional neural networks (CNNs) have become the dominant neural 
network architecture for solving many state-of-the-art (SOA) visual processing tasks. 
Even though Graphical Processing Units (GPUs) are most often used in 
training and deploying CNNs, their power efficiency is less than 10\,GOp/s/W for single-frame runtime inference. We propose a flexible and efficient CNN accelerator architecture called NullHop that implements SOA CNNs useful for low-power and low-latency application scenarios. NullHop exploits the sparsity of neuron activations in CNNs to accelerate the computation and reduce memory requirements. The flexible architecture allows high utilization of available computing resources across kernel sizes 
ranging from 1x1 to 7x7. NullHop can process up to 128 input and 128 output feature maps per layer in a single pass. 
We implemented the proposed architecture on a Xilinx Zynq
FPGA platform and present results showing how our implementation reduces external memory transfers and compute time in five different CNNs ranging from small ones up to the  widely known large VGG16 and VGG19 CNNs. Post-synthesis simulations using Mentor Modelsim in a 28nm process with a clock frequency of 500\,MHz show that the VGG19 network achieves over 450\,GOp/s. 
By exploiting sparsity, NullHop achieves an efficiency of 368\%, maintains over 98\% utilization of the MAC units, and achieves a power efficiency of over 3\,TOp/s/W in a core area of 6.3\,mm$^2$. As further proof of NullHop's usability, we interfaced its FPGA implementation with a neuromorphic event camera for real time interactive demonstrations.


\end{abstract}

\IEEEpeerreviewmaketitle

\section{Introduction}
 
Convolutional neural networks (CNNs) 
have emerged as one of the most popular approaches 
for solving a variety of large-scale machine vision 
tasks~\cite{Krizhevsky_etal12,He_etal15a,LeCun_etal15}. 
The conceptual simplicity of CNNs coupled with powerful supervised 
training techniques have made them the method of choice 
for extracting high-level semantic image features that 
form the basis for solving visual processing tasks such as classification, localization, and detection.%~\cite{Sermanet_etal13}.

CNN are trained, typically using backpropagation, to produce the correct output for a set of labeled examples. 
The network training is usually done on hardware platforms such as
graphical processing units (GPUs) or highly-specialized server-oriented architectures as in~\cite{Jouppi2017}. 

%After training, the CNN is used for runtime inference.
Inference in state-of-art (\textbf{SOA}) trained CNNs is computationally 
expensive, typically requiring several billion 
multiply-accumulate (\textbf{MAC}) operations per image. 
Using a mobile processor or mobile GPU to run 
%the inference step on a CNN 
inference on a CNN 
can become unfeasibly 
expensive in a power-constrained mobile platform. 
For example, the NVIDIA Tegra X1 GPU platform 
which targets mobile automatic driver assistance (ADAS) 
applications, can 
process 640x360 color input frames at a rate of 15Hz 
through a computationally efficient semantic 
segmentation CNN~\cite{paszke_enet:_2016}. 
Processing each frame through this CNN requires about 
2~billion MAC operations, thus the GPU does 
around 60 billion operations per second (GOp/s), 
at a power consumption of about 10W. 
Therefore at the application level, 
this GPU achieves a power efficiency of about 
6\,GOp/W, which is only about 6\% of its theoretical 
maximum performance. 
As a result, the NVIDIA solution can process a 
CNN at only 30 frames per second (FPS) if the 
network requires less than 2\,GOp/frame.

An important development in CNN research relevant 
to hardware accelerators are methods for training CNNs 
that use low precision weights, activations, and sometimes backpropagated gradients~\cite{Courbariaux15,Rastegari_etal16,Zhou_etal16, Courbariaux16,lin_fixed_2015,mishra_wrpn:_2017}. 
Training a network which has low-precision parameters 
and which uses the Rectified Linear Unit (ReLU) 
activation function leads  
up to 50\% increased sparsity in the activations. 
Several reported dedicated accelerators already exploit this sparsity~\cite{Chen_etal16,Moons2016,Moons2017}.

Because sparse networks can be beneficial for saving computes and memory access during inference, we 
developed a CNN hardware accelerator called \textit{NullHop} 
that exploits activation sparsity by two main features: 
The first feature is its ability to skip over 
zeros (zero-skipping) in the input CNN layers 
without any wasted clock cycles and redundant MACs. 
This method is different from~\cite{Han2016}, where a zero-skipping pattern 
is applied only in the computation of Fully-Connected (\textbf{FC}) layers. 
It is also different from~\cite{Zhang2016} 
where the convolutions are accelerated by exploiting 
the sparsity in the convolution kernels rather than 
in the activations. The second feature is a novel 
compression scheme that is optimized for sparsely 
activated CNN layers. This compression reduces 
external memory access and is more efficient 
than current run-length encoding schemes~\cite{Chen_etal16}.
It is also operates directly on compressed 
representations, unlike the work of~\cite{Moons2016}. 
The processing pipeline operates directly on the 
compressed input representation therefore more 
input data can be stored in the accelerator memory.
Finally, similar to the current SOA accelerators~\cite{pham_neuflow:_2012,Du2015,Chen_etal16,Sim_etal16,Moons2016,Moons2017,Lee18,Yin2017,Shin2017}, NullHop uses a configurable processing
pipeline that maintains high efficiency across a range of CNN kernel sizes and numbers of feature maps.

\section{CNN principles of operation}
\label{sec:cnns}
\begin{figure}[t]
  \centering
  \includegraphics[width=0.52\textwidth,trim=1.5cm -1cm -1cm 0]{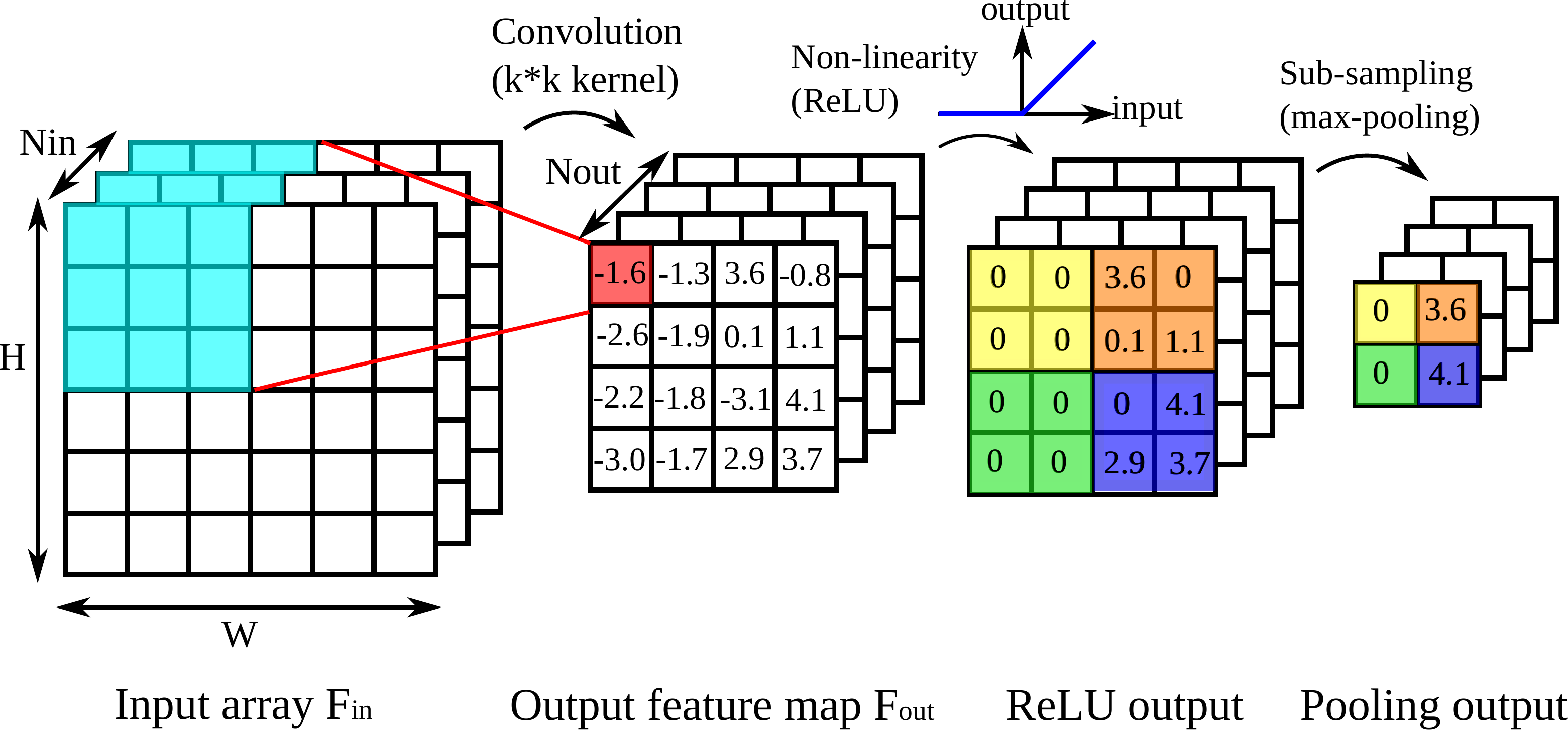}
  \caption{The three main processing stages in a CNN.}
\label{fig:conv_stages}
\end{figure}

CNNs extract high-level features from input images using successive stages of convolutions, non-linearities, and subsampling operations. The three stages are shown in Fig.~\ref{fig:conv_stages}. 
A typical vision application starts with 3 input feature map channels, corresponding to the red, green, and blue color channels of the camera image. 
The convolution stage takes then as input a 3D array ${\bf F^{in}}$ 
with $N_{in}$ 2D feature maps of size $H\times W$.
Each filter in the filter bank,
$c_{ij}$, is of size $k_w \times k_h$ and connects an input feature map $F^{in}_{i}$ to an
output feature map $F^{out}_{j}$ %which is defined 
as follows: 
\begin{equation}
F^{out}_j = \sum\limits_{\tiny i=0}^{N_{in}-1} c_{i,j}* F^{in}_i
\label{eq:conv}
\end{equation}

The convolution output is also a 3D array, ${\bf F^{out}}$  composed of
$N_{out}$ feature maps of size $N_{out}\times(H-k_h+1)\times(W-k_w+1)$.
A point-wise non-linearity is then applied to ${\bf F^{out}}$. In current SOA CNNs, the ReLU~\cite{Nair_Hinton10} is the most 
widely used non-linearity and is computed by $f(x) = max(0,x)$. 
In addition to being computationally cheap, using ReLUs empirically 
often yields better classification accuracy compared to saturating non-linearities, i.e, sigmoidal non-linearities~\cite{glorot_deep_2011}. 
The point-wise non-linear transformation is usually followed 
by a sub-sampling operation. A common sub-sampling strategy 
with many empirical advantages is max pooling~\cite{Scehere_etal10}, 
where each pooling window is replaced by the maximum value in the window. 
An example of a 2x2 non-overlapping pooling stage is shown in Fig.~\ref{fig:conv_stages}.

\subsection{Reduced Precision CNNs}
\label{sec:adaption}

When designing CNNs hardware accelerators, 
one major consideration is the energy consumption
from the amount of memory access and the number
of computes needed. Rounding a full precision
pre-trained network to lower precision weights and activations will help to
reduce hardware resources but the accuracy of
the network is usually compromised. It is
possible to increase the accuracy by also training 
deep networks using reduced bit precision methods as demonstrated in~\cite{stromatias_robustness_2015,Courbariaux15,Courbariaux16}.

\begin{figure}[!h]
  \centering
  \includegraphics[width=0.45\textwidth]{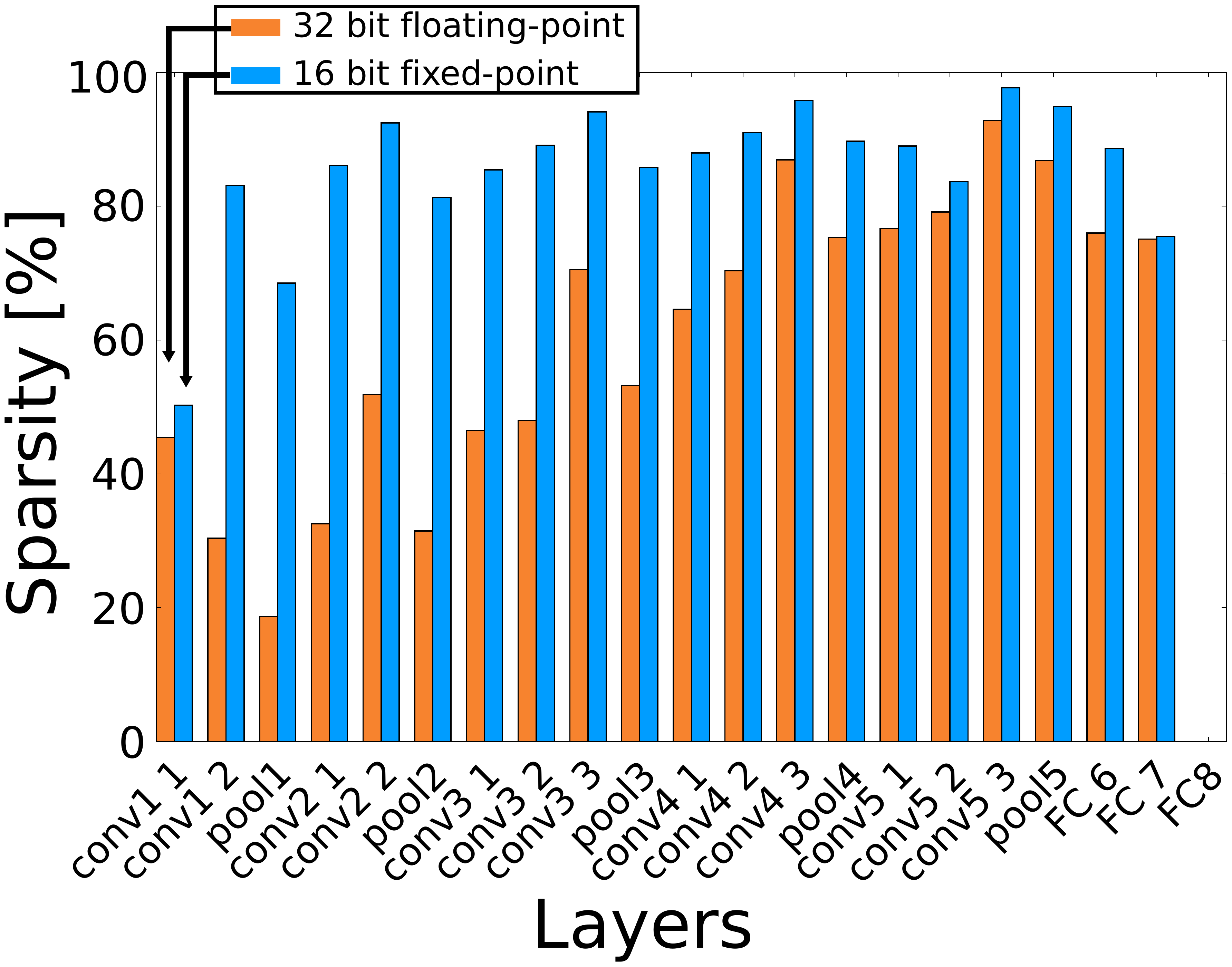} 
   \caption{Sparsity before (orange) and after (blue) activations are quantized  to 16-bit fixed-point for VGG16 layers. Average over 1000 ImageNet images.}
\label{fig:sparsity_comp}
\end{figure}

In order to run reduced precision CNNs on the NullHop accelerator, 
we developed a custom branch of Caffe\footnote{https://github.com/NeuromorphicProcessorProject/ADaPTION} called ADaPTION~\cite{milde_adaption:_2017}.
We use it to train networks from scratch as well as to fine-tune existing 32-bit floating-point networks to any specified fixed 
point precision for both weights and activations 
using the power2quant algorithm~\cite{stromatias_robustness_2015}. 
It also includes tools for estimating the required per-layer 
decimal point locations. We achieved VGG16 %~\cite{Simonyan_Zisserman14} 
67.5\% Top-1 accuracy after quantizing the weights and 
activations to 16 bits, with an accuracy drop of only 
0.8\% compared with the floating point VGG16. Reducing 
the number of bits also results in up 
to 50\% increased sparsity of activations per layer
as shown in Fig \ref{fig:sparsity_comp}. 
The average sparsity in the activations increased 
from 57\% in floating precision to a remarkable 82\% 
in reduced precision. An existing library in
Caffe called Ristretto~\cite{gysel2016hardware} can be used to train and test CNNs with reduced 
precision weights and activations but the 
weights are quantized only using a fixed-point 
notation, whereas the activations are quantized 
using 16-bit floating-point notation. 
Activations are much harder to represent 
using fixed-point, since the dynamic 
range of activations span nine orders 
of magnitude, for example, in the case of VGG16.
While others have investigated ultra low-precision networks (e.g. binary), these networks are 10X to 100X slower 
to train and require even more %multiple times as many 
feature maps to achieve the same accuracy~\cite{Courbariaux16,lin_fixed_2015,mishra_wrpn:_2017}. 
Therefore, we targeted a more generally useful 
hardware solution by quantizing weights 
and activations to an intermediate 16-bit fixed-point precision.

\section{Accelerator Architecture}

\begin{figure*}[h]
  \centering
  \includegraphics[width=0.7\textwidth]{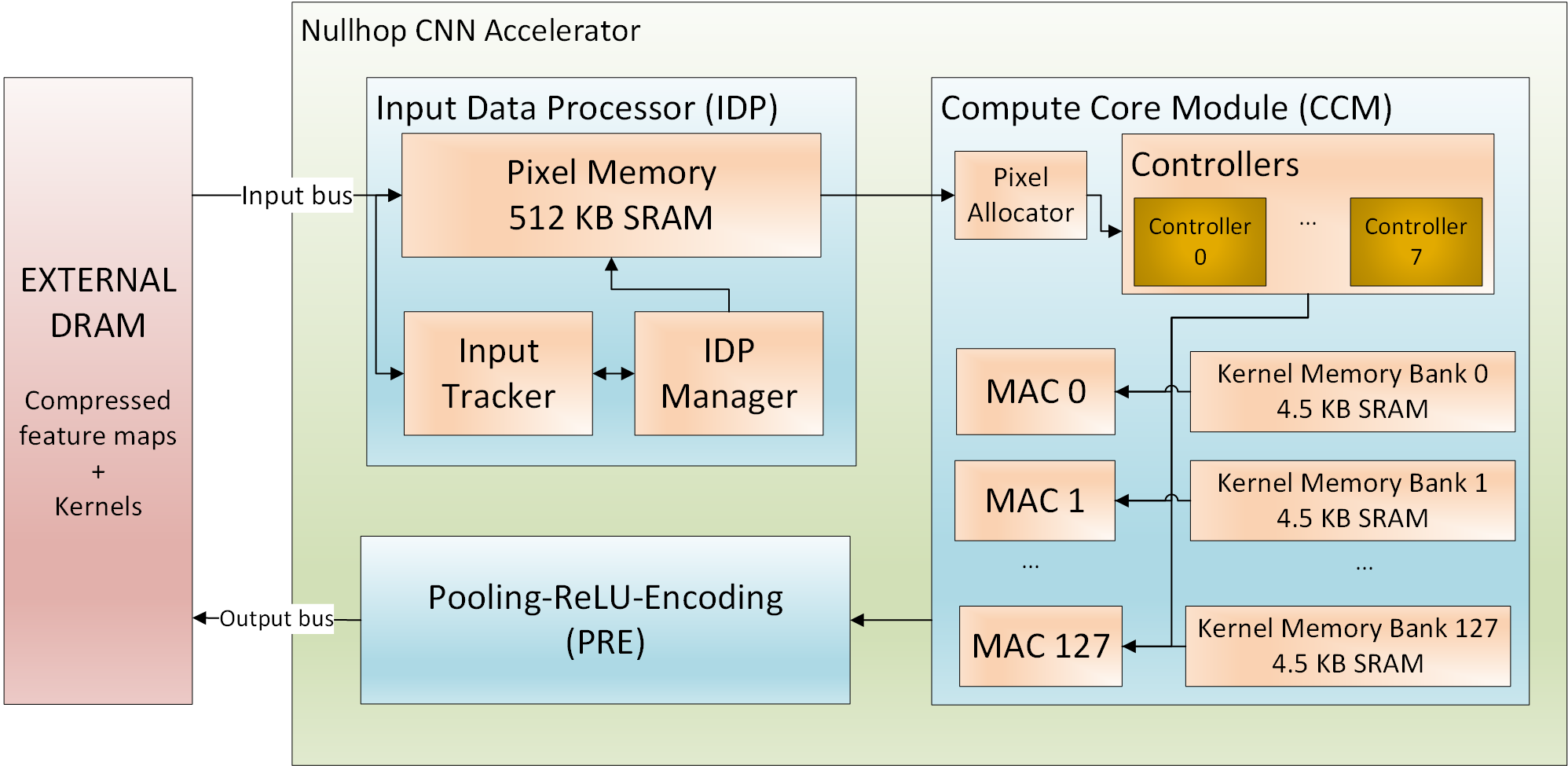} 
  \caption{High-level schematic of the proposed NullHop CNN accelerator.}
\label{fig:arch}
\end{figure*}

Figure~\ref{fig:arch} shows the high-level schematic of the NullHop accelerator. The accelerator interface is composed of two separate 32-bit data bus for the input and output, an input configuration interface that can be used by a host microcontroller for configuring the system, and a control interface for clock, reset and bus handshake signals. The accelerator implements one convolutional stage followed by a ReLU transformation and then a max-pooling stage. The ReLU and max-pooling stages can be disabled. To implement the full forward pass in a CNN, the accelerator evaluates convolutional stages one after another in a sequential manner. The input feature maps and the kernel values for the current convolutional layer are stored in two independent SRAM blocks. The internal memory structures are described in Sec.~\ref{sec:idp_decoding} and Sec.~\ref{sec:ccm}.

The output feature maps produced by the current layer are streamed off-chip to the external memory. They are then streamed back to the accelerator SRAM when the accelerator has finished processing the current layer. The feature maps are always stored in a compressed format that is never decompressed but rather decoded during the computation.

The following subsections describe the different functional blocks of the accelerator, the compression scheme employed, and the processing pipeline. We first give a high-level overview of the processing pipeline. The Input Decoding Processor (\textbf{IDP}) reads a small portion of the compressed input feature maps to generate pixels that are passed to the Compute Core Module (\textbf{CCM}). Only a few pixels, typically in one  mini-column (of height $k_h+1$) are fully decoded at any one time. These pixels are {\it all non-zero}. The input decoding block is able to directly skip over zero pixels in the compressed input feature maps without wasting any MAC operation. In addition to the pixel values, the input decoding block also forwards the pixels' positions (row, column, and input feature map index) to the CCM. The Pixel Allocator inside the CCM allocates each incoming pixel to a Controller. Each Controller manages the operation of a subset of the MAC blocks and submits the appropriate read requests to the Kernel Memory banks. All MAC blocks under the same Controller receive the same input pixel from their Controller, but weights from different kernels, producing pixels in different output feature maps. The convolution results are sent through an optional ReLU transformation and a max-pooling stage (implemented in the PRE, Pooling-ReLU-Encoding module) before going to the pixel stream compression block. The compressed output feature maps are then sent off-chip. 

\subsection{Sparse Matrix Compression Scheme}
\label{sec:compression}

\begin{figure}
  \centering
  \includegraphics[width=0.5\textwidth]{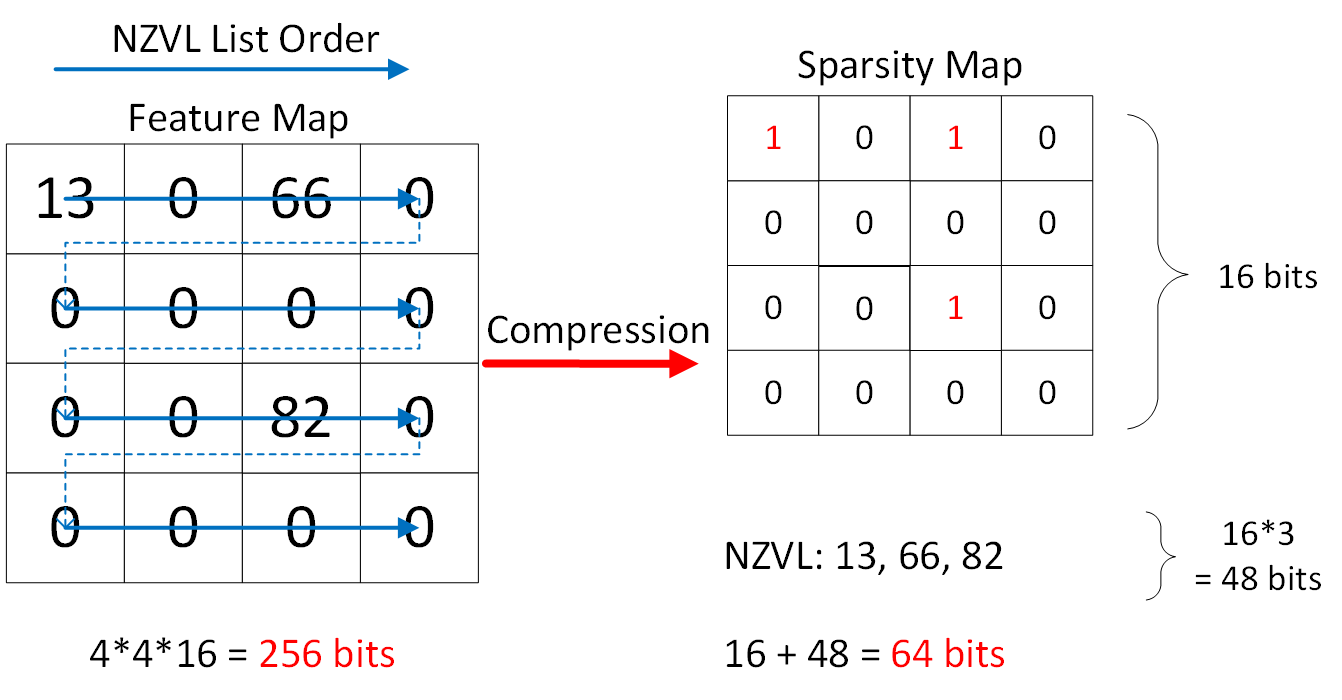} 
  \caption{Sparse compression scheme acting on a single sparse feature map. The non-zero values are stored as an ordered Non-zero Value List (NZVL). The order goes row-wise from top to bottom and from left to right in each row as shown on the feature map. The sparsity map (SM) is a binary mask with 1s at the positions of non-zero pixels, and 0s at the positions of zero pixels.}
\label{fig:compression_example}
\end{figure}

NullHop uses a novel sparse matrix compression algorithm. This algorithm produces an average compression level that is higher than that obtained from using the method in~\cite{Chen_etal16} and is easier to decode than the Huffman coding used in~\cite{Moons2016}.  
The coding uses two elements: a Sparsity Map (\textbf{SM}) and a Non-Zero Value List (\textbf{NZVL}). The SM is a 3D mask, having the same number of entries as number of pixels in the feature maps. Each binary entry in the SM is 1 if the corresponding pixel is non-zero, and 0 otherwise: 
\begin{equation}
SM(i, x, y)  =\begin{cases}1, input(i, x, y) \neq 0 \\0, {\rm otherwise}\end{cases} 
\end{equation}

The SM is used to reconstruct the positions of the non-zero pixels. These non-zero pixel values are stored as the NZVL: an ordered, variable-length list containing all the non-zero values in the feature maps. The compression scheme is illustrated in Fig.~\ref{fig:compression_example}. The accelerator sequentially reads both the SM and the NZVL and uses the information from the SM to decode the positions of the pixels in the NZVL in a sequential manner as described in the next subsection. The compressed image size (\textit{CIS}) in bits is given by: %(\ref{eq:compressed_size}):
\begin{equation}
CIS = E(n + N(1- S_p))
\label{eq:compressed_size}
\end{equation}
\noindent where
\vspace*{1\baselineskip} %empty line
\begin{tabular}{@{}>{$}l<{$}l@{}}
CIS & input image size in bits\\
E & total number of pixels in input image\\
S_p & sparsity of input image (range 0-1)\\
N & input precision in bits\\
n & 1 bit
\end{tabular}
\vspace*{1\baselineskip} %empty line

\begin{figure}

\centering
\begin{subfigure}{0.35\textwidth}
\hspace*{-1cm} 
\includegraphics[height=0.9\textwidth, width=1.1\textwidth]{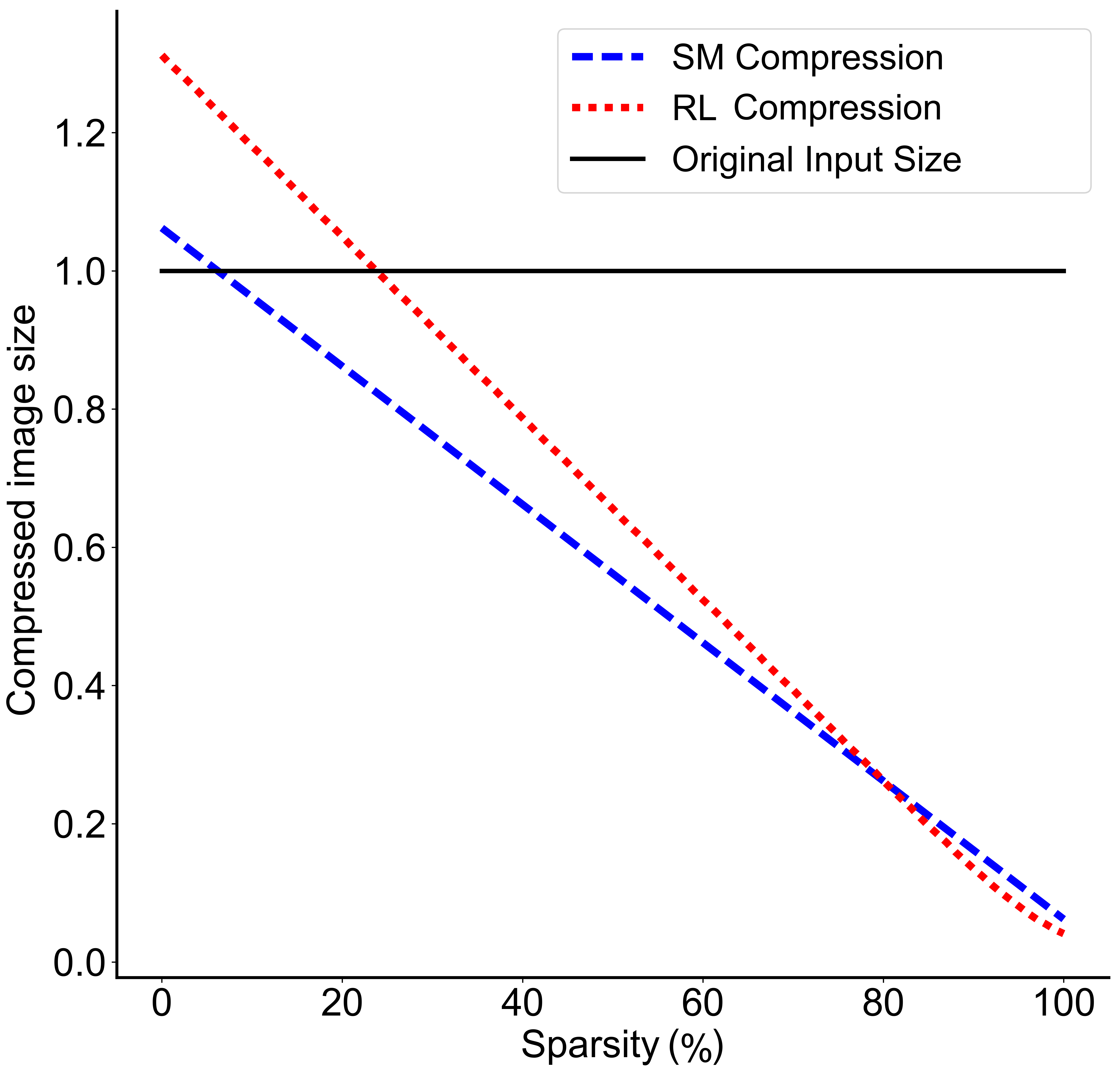}.
\caption{Comparison of compression methods over different sparsity amounts. Results from 10,000 images.}
\label{graph:sparsity_step}
\end{subfigure}

\begin{subfigure}{0.35\textwidth}
\hspace*{-1cm} 
\includegraphics[height=0.9\textwidth, width=1.1\textwidth]{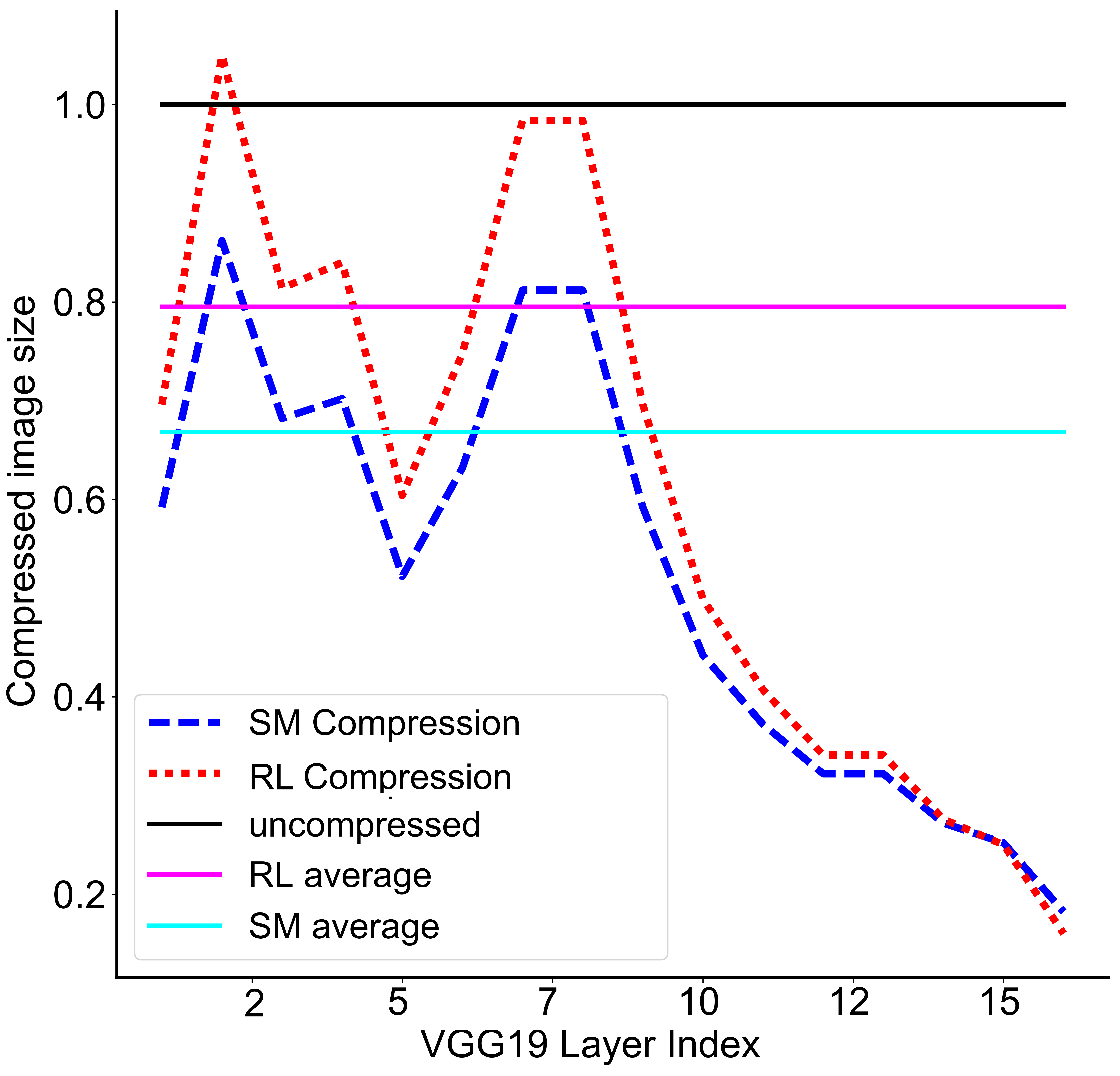}.
\caption{Comparison on 1000 runs of VGG19.}
\label{graph:vgg19_comparison}
\end{subfigure}

\caption{Compression performance comparison between the Sparsity Map (SM) and Run-Length (RL) Compression algorithms.}
\label{fig:sparsity_comparison}
\end{figure}

Eq.~\eqref{eq:compressed_size} shows how the size of the compressed input image has a lower limit \textit{CIS = E} when sparsity \textit{$S_p$ = 1}. From \eqref{eq:compressed_size}, it is also possible to demonstrate that the algorithm provides a reduction in memory when condition (\ref{eq:threshold_sparsity}) is satisfied: 
\begin{equation}
S_p > Th_p=1/N%\frac{1}{N}
\label{eq:threshold_sparsity}
\end{equation}
where $Th_p$ is defined as the threshold sparsity.
Solving \eqref{eq:threshold_sparsity} for different data bit precisions leads to Table~\ref{table:bit_precision} where a precision of 16 bits shows a threshold sparsity of 0.0625 which is low enough to guarantee compression in most CNN layers. Our compression algorithm shows better average compression performance than the run-length (RL) compression algorithm proposed by~\cite{Chen_etal16} %for almost all compression levels 
as demonstrated for the VGG19 example in Fig.~\ref{fig:sparsity_comparison}. In Fig.~\ref{graph:vgg19_comparison} it is also possible to see how the RL algorithm cannot reduce the size of the output feature maps for some layers, effectively increasing their size. Although our sparsity map algorithm produces 
an equivalent level of compression as the Huffman coding used by~\cite{Moons2016}, our data structure permits an easier decoding during the computation, allowing the accelerator to operate directly on compressed representations without decompression.

\begin{table}[h]
\centering
\caption {Minimum sparsity needed for compression of input feature maps.}
\begin{tabular}{l*{2}{c}r}
Bit Precision    &  Threshold Sparsity\\
\hline
8 & 0.1250 \\ 
12 & 0.0833 \\
16 & 0.0625 \\
24 & 0.0416 \\
32 & 0.0312 \\
\end{tabular}
\label{table:bit_precision}
\end{table}

\subsection{Pixel Memory and Decoding: Input Data Processing Unit}

\subsubsection{Input format}
 
The 3D SM is split into 16-bit word segments that are streamed to the CNN accelerator interleaved with the corresponding NZVL pixels. The length of the SM segments is implementation dependent and in our case, it is equivalent to the chosen bit precision of the activation to simplify the hardware implementation.

\begin{figure}
  \centering
  \includegraphics[width=0.35\textwidth]{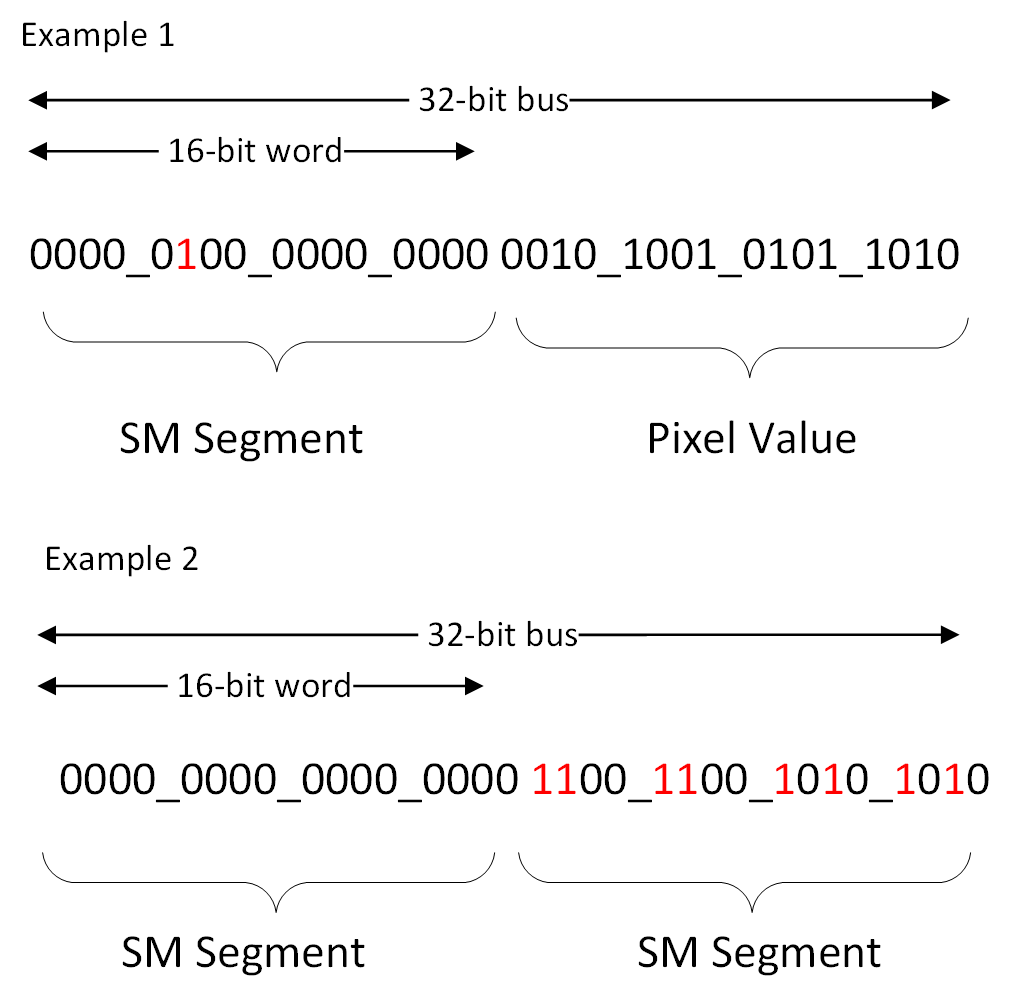} 
\caption{Format of the words sent to the accelerator, using 16-bit words and 32-bit bus as an example. 16-bit SMs are interleaved with non-zero pixel values. If a SM has all-zero entries, it is followed by next SM.}
\label{fig:pixel_sm}
\end{figure}

The number of ones in a SM segment indicates the number of pixel values that will follow the SM segment as shown in Example 1 of Fig.~\ref{fig:pixel_sm}. The position of the ones indicates the offsets of these pixels. The first word that is sent to the accelerator is always a SM segment. All fields afterwards can be SM segments or pixel values. If there is a run of 16 zero pixels, a SM segment can be all zeros as shown in Example 2 of Fig.~\ref{fig:pixel_sm}. The SM segment will then be followed by another SM segment and this will continue until a SM segment that is not all zeros is streamed in. 

The compressed rows are streamed into the accelerator one after the other starting from the top row: let $p(i,x,y)$ be the pixel at position $(x,y)$ of the $i^{th}$ input feature map, $F^{in}_i$. The pixels are streamed into the accelerator as follows:
$p(0,0,0),p(1,0,0),..,p(N_{i},0,0),p(0,1,0),..,p(N_{i},1,0),..$ 

\subsubsection{Decoding the compressed rows}
\label{sec:idp_decoding}
The IDP is responsible for storing and decoding the compressed rows of the input feature maps. The module contains multiple SRAM banks and can start decoding the data from these banks while the input feature maps are still being loaded.  The IDP maintains a pointer to the beginning of each row of the image stored inside the SRAM and uses these row starting addresses to decode the pixels in a sequential manner. 

At each clock cycle, the IDP can read up to $k_h +1$ non-zero pixels from memory to be sent to the CCM module (Fig.~\ref{fig:arch}). The pixels are read in a winding manner within a vertical stripe as shown in Fig.~\ref{fig:conv}. 
The pixels within this stripe are the only pixels needed to generate a double row in the output feature maps, assuming a vertical convolution stride of 1.  The IDP supports zero padding at feature map boundaries without having to load any extra data and without wasting any clock cycles. This is possible by adding proper offsets to each pixel position while sending them to CCM module.

There are three main blocks inside the IDP module:
\paragraph{Pixel Memory}
The Pixel Memory block stores the input feature maps to be processed and is composed of several 
SRAM memory banks and arbiters. The arbitration scheme gives maximum priority to write requests from off-chip memory, followed by read requests received by the IDP Manager (described in Sec.~\ref{par:idp_manager}). The block also handles the communication between the IDP and CCM modules, sending pixel values to the processing unit and stalling IDP operations if CCM cannot receive more data. 

\paragraph{Input Tracker}
In order to access rows in compressed mode, a memory that stores the pointers to each row's starting position in the Pixel Memory is required. The Input Tracker accomplishes this task by storing starting position addresses received by the Pixel Memory in a small SRAM memory when a new row is stored; and providing them to the IDP Manager when requested. The read/write arbitration in the Input Tracker follows the same scheme as the one in the Pixel Memory.

\paragraph{IDP Manager} 
\label{par:idp_manager}
IDP operations are controlled by the IDP Manager, which consists of a set of FSMs acting as control units for the IDP module. The number of FSMs is equal to the maximum kernel size supported plus 1; each control unit sends read requests to Pixel Memory for a specific row in the currently active (vertical) stripe. During processing, $k_h+1$ FSMs are enabled, while others are disabled. For example, for $k_h=3$ there are $3+1=4$ enabled FSMs, each one reading a different row of the input. Each FSM stores in its internal registers, a SM and a memory pointer to the next address to be read, plus a register containing the spatial coordinates of the next pixel to be read. 

At the beginning of the layer, the IDP Manager reads the pointers to the first $k_h+1$ rows from the Input Tracker and stores them into each IDP FSM memory pointer register. 
The FSMs then send a read request to the Pixel Memory and store the read result into their SM registers.
In the same clock cycle, the position of the first non-zero entry is computed using the SM itself and the memory pointer register is incremented by one before the next read operation. 
%to be ready for the next read operation. 
Next, clock cycles are dedicated to the actual reading of pixels from memory: 
The FSMs first send a read request to the Pixel Memory for the current memory pointer (providing its spatial coordinates as extra information to be forwarded to CCM). They then increment the memory pointer by 1, and look up the next non-zero entry in the SM in order to get the coordinates of the pixel to be read in the next clock cycle. 
The IDP module iterates over SMs and pixels until the row ends, moving to the next feature map stripe when all FSMs have completed their task. No pixel with zero value is forwarded to the CCM module.

\subsection{Compute Core Module: Pixel Allocator, Controllers, and MAC Blocks}
\label{sec:ccm}

\begin{figure}[t]
  \centering
  \begin{subfigure}[b]{0.48\textwidth}
    \includegraphics[width=\textwidth]{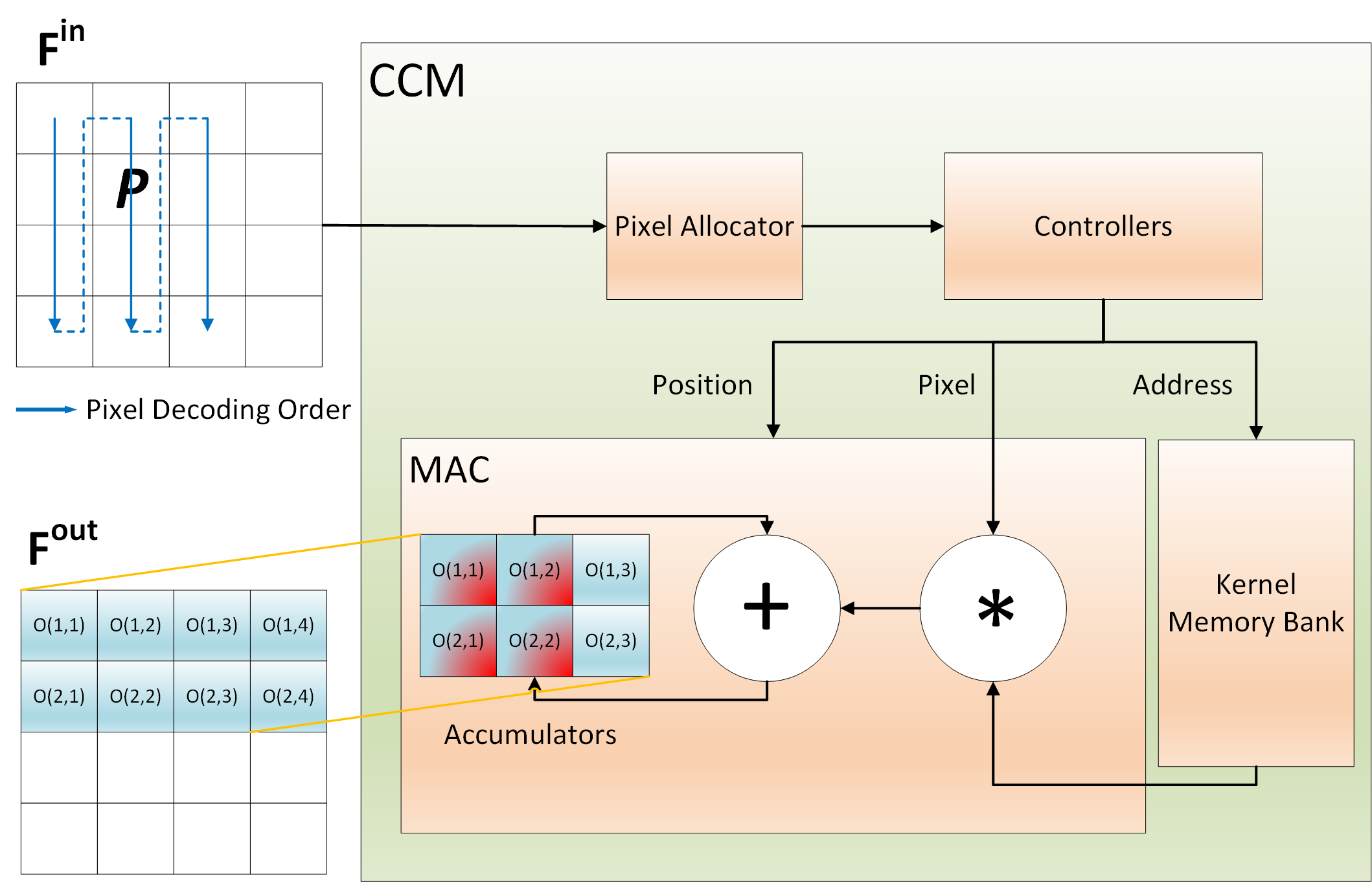} 
    \subcaption{} 
    \label{fig:conv_a}
  \end{subfigure}
  \quad
  \begin{subfigure}[b]{0.48\textwidth}
    \includegraphics[width=\textwidth]{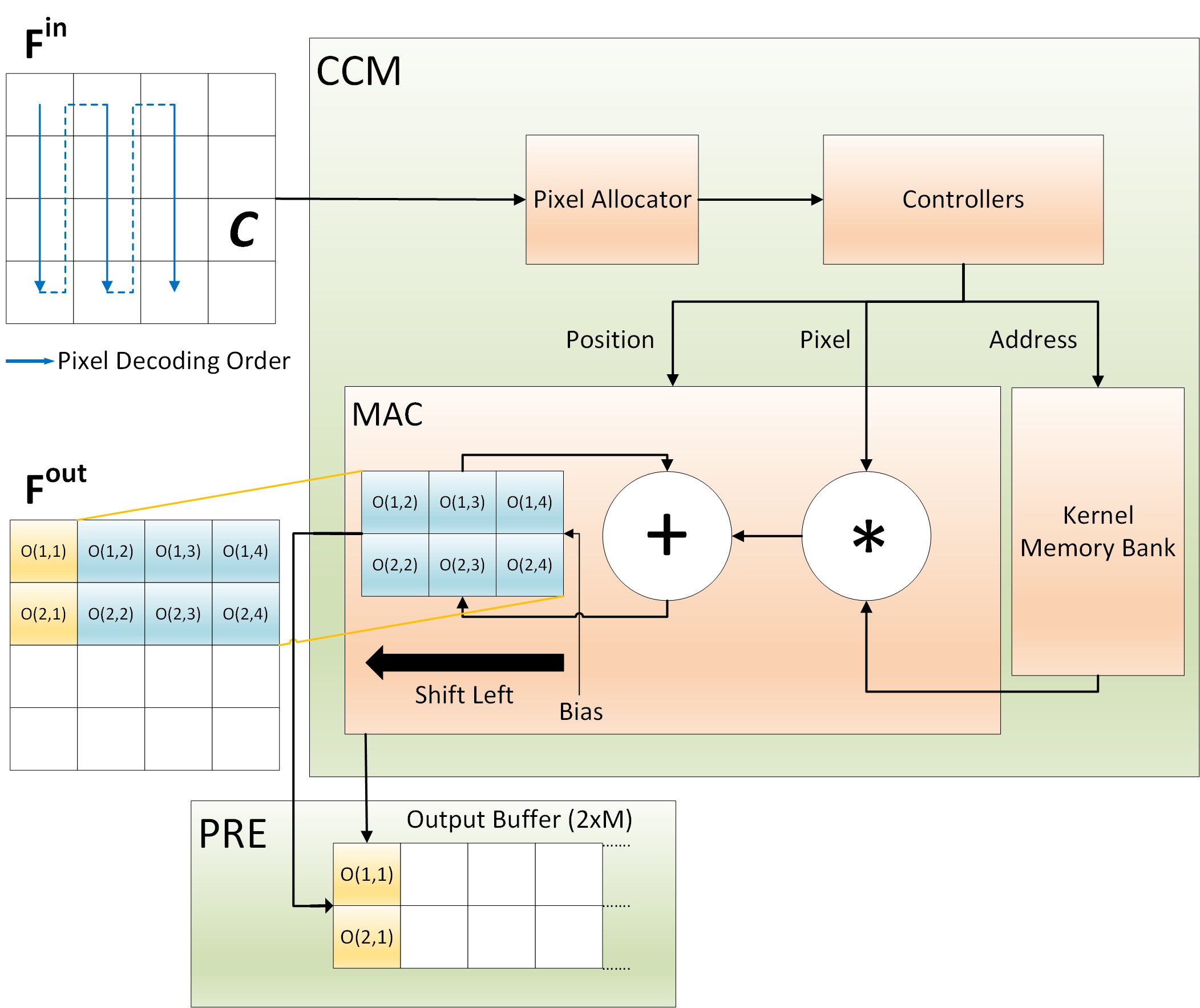} 
    \subcaption{} 
    \label{fig:conv_b}
  \end{subfigure}
\caption{Pixel processing scheme for an example case of a $3\times 3$ kernel, one input feature map, and $128$ output feature maps. (\subref{fig:conv_a}) One MAC block is shown, the pixel $P$ is used to update the red-shaded results in the current output patch. (Sec. \subref{fig:conv_b}) On receiving pixel $C$, the controller detects (based on the column index of $C$) that the computation of output pixels stored in the most-left entries of the accumulator array (pixels $o(1,1)$ and $o(2,1)$) is complete. Pixels $o(1,1)$ and $o(2,1)$ are shifted into an output buffer.
}
\label{fig:conv}
\end{figure}

The CCM is the arithmetic unit of the architecture and computes the convolution output ${\bf F^{out}}$. It is composed of the following blocks:

\begin{itemize}
\item[-] A $Kernel$ $Memory$ with $M$ SRAM banks. It stores the kernel weights necessary to compute the current layer. 

\item[-] A $MAC$ $block$ containing $M$ MACs, each composed of a multiplier, an adder, and $2 \times k_{w}$ accumulators. 

\item[-] A $block$ of $C$ $Controllers$, each overseeing the weights read from the Kernel Memory and redirecting the weights and activations to a subset of MACs called $cluster$.

\item[-] A $Pixel$ $Allocator$ that oversees the redirection of the pixels received from the IDP to the different controllers.
\end{itemize}

\paragraph{Basic Operations}
We now describe the simplest accelerator configuration where $N_{out} = M$. At the beginning of a layer computation, the weights are loaded in the Kernel Memory. Each kernel is stored in a different Kernel Memory Bank and assigned to a MAC block, and the size of the only existing cluster is $M$. The Pixel Allocator forwards pixels from the IDP to the only active Controller. At every clock cycle, the Controller sends a single address, based on the coordinates $x$ and $y$ of the pixel itself, to all the Kernel Memory banks. Consequently, all MACs receive the same pixel coordinate and weights from different kernels, each computing the pixel value for a different output channel.

\paragraph{Kernel Read Scheme}

NullHop computes ${\bf F^{out}}$ in a row-wise manner, with 2 output rows computed at the same time to enable on-fly 2x2 pooling. In a single MAC, pixel $P$ is multiplied by all the weights of two rows of a kernel. Each multiplication result is added up in a different entry of the MAC accumulators according to the coordinates of $P$ and of the weights. Fig.~\ref{fig:conv_a} illustrates this process for the case of 1 input feature map and when $k_w=k_h=3$. $P$ is multiplied by the weights in position $k_{0,0}$, $k_{0,1}$, $k_{1,0}$, $k_{1,1}$. Since it is close to the left image boundary, it is not necessary to multiply it with weights in position $k_{0,2}$ and $k_{1,2}$, since these multiplications do not contribute to any value of ${\bf F^{out}}$.

Fig.~\ref{fig:conv_b} shows what happens when a pixel lies in a different column than the previous one. In this example, the Controller detects that there is a column index increase. Because the IDP sends the pixel along vertical stripes, the values stored in the left-most entries of the MAC accumulators are the convolution results for the previous column. The Controller asserts a shift flag to the MAC accumulators, which are shifted left and sent to the output buffer inside the PRE module (described in Sec.~\ref{subsection:pre}). The right-most entries of the accumulators are initialized with the value of the bias, ready for the computation of next output column. This process repeats: As the input decoding moves to a new column, the accumulators are shifted one position to the left until all the pixels in a double row of ${\bf F^{out}}$ are produced.

\paragraph{Different Input Feature Maps}
\label{sec:multipleinputfeaturemaps}
When the number of output feature maps of a layer is different from $M$, the function of the Pixel Allocator and the Controllers changes to redistribute the workload over multiple MACs.

1)	$N_{out} < M$: Since there are more MACs than output feature maps, it is impossible to perform a 1-to-1 mapping. Instead, NullHop splits the computation of a single output channel over $ v= M \div N_{out}$ MACs.  The Pixel Allocator sends, in parallel, pixels to $v$ Controllers. Each controller is in charge of issuing commands to a $cluster$ composed of $N_{out}$ MACs and a Kernel Memory Bank. Every MAC stores partial convolution results that will be summed up in the PRE module to produce ${\bf F^{out}}$.

2)	$N_{out} > M$: In this case, it is impossible to assign an output feature map to every MAC block. To handle this situation, NullHop splits the computation over multiple passes. In each pass, we process a subset of the output feature maps. First, we load the first $M$ kernels and compute $F^{out}_{(0:M-1)}$. Then, we load a second set of $M$ kernels and compute $F^{out}_{(2M:2M-1)}$. The procedure is repeated until all the $N_{out}$ channels are processed. If $N_{out}$ is not a multiple of $M$, the system distributes the kernels evenly among passes and computes them as described for $N_{out} < M$.

3) $N_{out}$ $[KBytes]$ $>$ $Kernel$ $Memory$ $Bank$ $[KBytes]$: In some cases, it is possible that $M$ kernels cannot be allocated to $M$ memory banks because they would not fit in the SRAM banks (e.g. in case of 7x7 convolutions with $M$ input channels). Instead of increasing the memory size to cover also these corner cases, NullHop splits the kernels over multiple memory banks and passes similar to what was described in 1) and 2).

%%%%%%%%

\subsection{Pooling, ReLU, and Encoding Unit}
\label{subsection:pre}
The pooling, ReLU, and Encoding module (\textbf{PRE}) is responsible for the final processing steps of the computational pipeline. It receives the convolution results shifted out from the MACs in the CMM 
and stores them in the PRE buffer. In the current implementation, the buffer size is equal to $2 * M$ memory entries where 2 is the pooling dimension. The ReLU non-linearity can be applied and $2\times 2$ pooling can be performed on the fly. An encoder is used to compress the output pixels according to the scheme described in Sec.~\ref{sec:compression} and to stream out the SM segments and non-zero pixels.

If the number of MAC blocks per cluster is larger than 1, the MAC blocks produce partial convolution results. Before the data can be further processed, these partial results need to be summed to produce the {${\bf F^{out}}$ values. The accumulation is performed within the PRE buffer, in $log(size(cluster))+1$ clock cycles. During each accumulation cycle, two adjacent partial values are summed together and the result is stored back in the PRE buffer. Once the accumulation is over, the results are transferred to the output buffer which can store up to $M$ pixels.

The ReLU non-linearity is applied during the transfer of pixels from the PRE buffer to the output buffer. Each entry of the output buffer is initialized to zero if the ReLU non-linearity is enabled, otherwise with the most negative number that can be represented. When transferring pixels from the PRE buffer to the output buffer, a max operation is performed so that the value stored in the output buffer is the maximum of the incoming pixel and the original stored value. 
When pooling is enabled, the maximum of the three values of the pixel pair and the current content of the output buffer is stored back into the output buffer.

When the second pixel pair arrives from the CCM, the same max operation is carried out. In this way the output buffer contains the results of the $2\times 2$ max-pooling.
When pooling is disabled, one row at a time is transferred from the PRE buffer to the output buffer. The encoder acts on the first $B$ pixels of the output buffer, a number of pixels equal to the bit precision of the chip. In our 16-bit implementation, the encoder works on 16 pixels at a time. The encoder generates a SM segment from the first $B$ pixels based on the position of the non-zero pixels. For each set of $B$ pixels, the encoder streams out the SM and the first non-zero pixel during the first clock cycle; then, two pixels are sent out on every subsequent clock cycle. 
When all the pixels of the current SM have been streamed out, the output buffer is shifted by $B$ positions, a new SM is generated and the non-zero pixels streamed out. This is repeated until all the pixels in the output buffer are processed.
The encoding scheme can be switched off. In this case, all pixels are streamed out and no SM is generated. This option allows the host software to obtain the output activation in a easily accessible format, speeding up any eventual extra processing the CNN may require.

\section{Design Implementation}
The NullHop architecture was implemented using the following parameters:
\begin{itemize}
  \setlength\itemsep{0em}
\item 16-bit precision fixed point kernels/activations
\item 32-bit precision MAC units
\item 32-bit input/output bus
\item IDP memory: 512 KB
\item Kernel memory: 576 KB
\item Number of MAC: 128
\item Number of controllers: 8
\item Max supported kernel size: 7
\item Max number of rows in input image\footnote{\label{fn:counter}Determined by counter resolution; chosen for tested networks. No effect on throughput or area.}: 512 

\item Max number of columns in input image\footnotemark[\value{footnote}]: 512
\item Max number of feature maps\footnotemark[\value{footnote}]: 1024 

\end{itemize}

\subsection{Synthesis, Place and Route}
We synthesized NullHop with Synopsys Design Compiler using Globalfoundries 28nm technology. From this manufacturing process, we used the library with the smallest available standard cells, including all the different threshold voltages except for the lowest. This choice was to reduce area and power for embedded system deployment. Fig.~\ref{fig:placeroute} shows the results of the place and route, including the position of the I/O pads. The post place-and-route core size is 6.3\,mm$^2$ (core utilization = 70\%) and the chip size including I/O pads is 8.1\,mm$^2$, with an estimated power consumption of $155$\,mW at 1\,V supply voltage and a clock frequency of 500\,MHz. The power consumption was estimated using switching activity with Synopsys Design Compiler. The result allows us to estimate the compute performance per Watt of the accelerator as reported in Table~\ref{table:result_comparison}. 

\begin{figure}[tb]
  \centering
  \includegraphics[width=0.35\textwidth, height=0.325\textwidth]{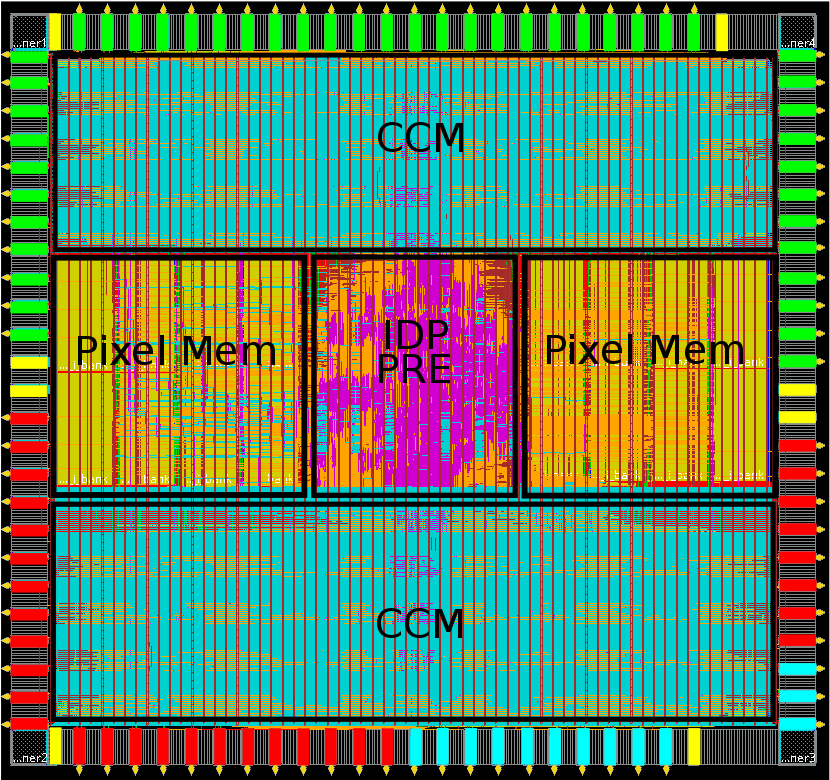} %nhp_3}
  \caption{NullHop chip place and route. Pads for power (yellow), input data and input configuration bus (green), output data bus (red) and control signals (blue) are highlighted.}
\label{fig:placeroute}
\end{figure}

\subsection{FPGA Implementation}
\label{sec:fpga}

To validate the design, we implemented it on a Xilinx Zynq 7100 System-On-Chip (\textbf{SoC}) FPGA, using the AXI4-Stream with Direct Memory Access
(\textbf{DMA}) open source protocol to connect the SoC FPGA fabric with its
ARM processor. The ARM CPU runs Petalinux as the operating system (OS) to control
the accelerator. It manages the read and write operations between the DDR
memory of the ARM computer and the BRAM of the accelerator. The processor also
computes the FC layers following the convolutional layers. We included in the
OS an embedded USB host controller module to interface it to an iniLabs
DAVIS240C neuromorphic event-based camera~\cite{Brandli_etal14} for the real-
time demonstrations described in Sec.~\ref{sec:application_example}. For this
use, the ARM processor also runs iniLabs
cAER\footnote{https://inilabs.com/support/software/caer/}, an open source
framework to interface to the DAVIS camera. Fig~\ref{fig:ARMAXINullHop} shows
the block diagram of our FPGA architecture, including the MM2S ($Memory$ $To$ $Stream$) and S2MM ($Stream$ $to$ $Memory$) modules used to interface the accelerator with the AXI4-S bus.

The design minimizes host memory manipulation by using DMA transfers from host
memory to the accelerator and vice versa without requiring each layer output to be reformated or processed. For each layer, the ARM loads the layer configuration and
the kernels. It then initiates interrupt-driven DMA transfer of the input and output. It is then free for other processing while the layer is computed. The development of the complete functionality required at least 12 months of effort beyond the basic reference IP from Xilinx.

\begin{figure}[t]
  \centering
  \includegraphics[width=0.45\textwidth]{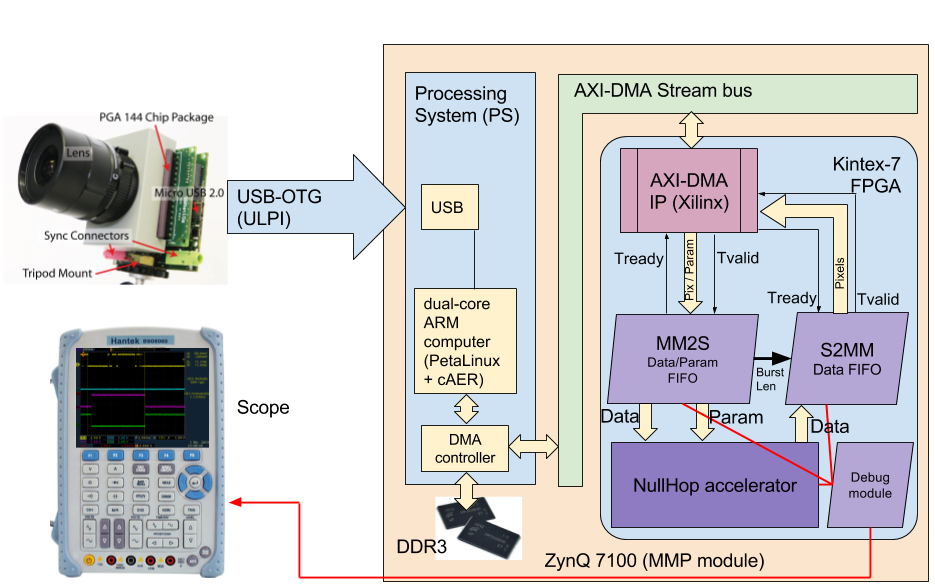} 
  \caption{System-on-Chip (SoC) block diagram for the testing scenario. The FPGA on the Zynq SoC hosts the NullHop accelerator plus glue logic for interfacing to the Zynq ARM processor DDR memory using the DMA.}
\label{fig:ARMAXINullHop}
\end{figure}

Our IC implementation targets a clock frequency above 500\,MHz, but routing
limits the FPGA implementation to much lower frequencies. For the Zynq 7100,
the maximum clock frequency is 60\,MHz after synthesizing and implementing the
entire infrastructure which includes NullHop and the AXI interfaces. Table~\ref{table:FPGA-resources} shows the required resources.

\begin{table}[t]
\centering
\caption{Resources used on NullHop Zynq 7100. %hierarchical resource consumption. 
%\AAI{To be updated by Seville}\AL{Updated.}\AR{Updated after Tobi's recommendation}
}
\label{table:FPGA-resources}
\begin{tabular}{|l|c|c|c|c|}
\hline
Resources                                                                & Logic                                                     & FF                                                        & BRAM                                                     & DSP                                                   \\ \hline
\begin{tabular}[c]{@{}l@{}}Full System\end{tabular} & \begin{tabular}[c]{@{}c@{}}229K\\ (83\%)\end{tabular} & \begin{tabular}[c]{@{}c@{}}107k\\ (19\%)\end{tabular} & \begin{tabular}[c]{@{}c@{}}386\\ (51.1\%)\end{tabular} & \begin{tabular}[c]{@{}c@{}}128\\ (6.3\%)\end{tabular} \\ \hline
\end{tabular}
\end{table}

\begin{table}[b]
\centering
\caption{Power consumption estimation of NullHop on Zynq 7100.} 

\label{table:SoCPower}
\begin{tabular}{|l|c|c|c|c|c|c|c|}
\hline
\begin{tabular}[c]{@{}l@{}}Power \\ (mW)\end{tabular}    & \begin{tabular}[c]{@{}l@{}}Dynamic\end{tabular} & FPGA & Logic & BRAM & DSP & Routing \\ \hline
\begin{tabular}[c]{@{}l@{}}NullHop + \\ AXI4-s + ARM\end{tabular}       & 2339    & 804 & 20   & 396  & 2  & 67     \\ \hline
NullHop                                                              & 777    & 777 & 19   & 384  & 2  & 63     \\ \hline
IDP                                                                  & 97     & 97  & 2    & 75  & -   & 13      \\ \hline
\begin{tabular}[c]{@{}l@{}}CCM\end{tabular} & 638    & 638  & 13   & 309  & 2  & 43     \\ \hline
PRE                                                                  & 41     & 41   & 4     & -    & -   & 7       \\ \hline
\end{tabular}
\end{table}

Power analysis for the implemented NullHop on the Zynq 7100 was estimated
with Vivado assuming half the nodes switch state on each clock cycle for the logic.
These estimates show a static power consumption of 316\,mW when the FPGA clock is stopped and the ARM cores are idle; and a dynamic power consumption of 1.5\,W for the ARM
processor and 0.8\,W for NullHop plus the AXI4-S logic.

Table~\ref{table:SoCPower} shows the breakdown of the power estimate across the blocks. %distributed 
%among the NullHop infrastructure.
The ARM consumes 63\% of the total power, 
while the logic circuits and memory on the Kintex-7 embedded FPGA consume the remaining 37\%. The power breakdown for the FPGA blocks is estimated as 27.27\% for CCM, 4.14\% for IDP, 1.75\% for PRE, and 1.15\% for AXIstream. 
%where it is distributed between components 
%as 27.27\% for CCM, 4.14\% for IDP, 1.75\% for PRE, and  1.15\% for AXIstream. 
%The ARM consumes 63\% of the total power, 
%while 37\% is used by the logic circuits and memory in the Kintex-7 embedded FPGA, 
%where it is distributed between components 
%as 27.27\% for CCM, 4.14\% for IDP, 1.75\% for PRE, and  1.15\% for AXIstream. 

To validate the Xilinx estimations, we measured the power consumption of the different stages of the complete testing infrastructure when running the example described in Sec.~\ref{sec:roshambo}. The base-board that powers the AvNet 7100 SoC mini-module consumes 5.1\,W. After adding the 7100 and the fan, but with no design in the FPGA and in reset mode, it consumes 6.95\,W. After programming the FPGA, in an idle linux state not running cAER, the
system consumes 8.27\,W. Continuously running RoShamBo increases the power to 9\,W.
Thus the incremental power is about 750\,mW, close to Vivado's estimate of
777\,mW.

\section{Application example}
\label{sec:application_example}
We studied several CNNs, ranging from small 
custom CNNs on a classification task 
with a small number of labels to 
widely-used large CNNs used to classify 
ImageNet, a large dataset with 1000 classes.

\begin{table}[]
\centering
\caption{Face Detector: Parameters of the 2 convolutional layers}
\label{table:face_detector}
\begin{tabular}{|l|c|c|c|c|c|c|}
\hline
\begin{tabular}[c]{@{}l@{}}Layer\\ Number\end{tabular}                                                                
& \begin{tabular}[c]{@{}l@{}} Input \\ feature\\maps\end{tabular}                                                    
& \begin{tabular}[c]{@{}l@{}} Output \\feature\\maps\end{tabular} 
& \begin{tabular}[c]{@{}l@{}} Kernel\\Size \end{tabular}                  
& \begin{tabular}[c]{@{}l@{}} Input \\Width/Height\end{tabular}                                                     
& Pooling & \begin{tabular}[c]{@{}l@{}} Number\\of passes\end{tabular}                                                   
\\ \hline
1 & 1 & 16 & 5 & 36x36 & Yes & 1\\
2 & 16 & 16 & 3 & 16x16 & Yes & 1 \\
\hline
\end{tabular}
\end{table}

\begin{table}[]
\centering
\caption{RoshamboNet: Parameters of the 5 convolutional layers}
\label{table:roshambonet}
\begin{tabular}{|l|c|c|c|c|c|c|}
\hline
\begin{tabular}[c]{@{}l@{}}Layer\\ Number\end{tabular}                                                                
& \begin{tabular}[c]{@{}l@{}} Input \\ feature\\maps\end{tabular}                                                    
& \begin{tabular}[c]{@{}l@{}} Output \\feature\\maps\end{tabular} 
& \begin{tabular}[c]{@{}l@{}} Kernel\\Size \end{tabular}                  
& \begin{tabular}[c]{@{}l@{}} Input \\Width/Height\end{tabular}                                                     
& Pooling & \begin{tabular}[c]{@{}l@{}} Number\\of passes\end{tabular}                                                   
\\ \hline
1 & 1 & 16 & 5 & 64x64 & Yes & 1\\
2 & 16 & 32 & 3 & 30x30 & Yes & 1 \\
3 & 32 & 64 & 3 & 14x14 & Yes & 1\\
4 & 64 & 128 & 3 & 6x6 & Yes & 1\\
5 & 128 & 128 & 1 & 2x2 & Yes & 1\\ 
\hline
\end{tabular}
\end{table}

\begin{table}[]
\centering
\caption{Giga1Net: Parameters of the 11 convolutional layers}
\label{table:giganet}
\begin{tabular}{|l|c|c|c|c|c|c|}
\hline
\begin{tabular}[c]{@{}l@{}}Layer\\ Number\end{tabular}                                                                
& \begin{tabular}[c]{@{}l@{}} Input \\ feature\\maps\end{tabular}                                                    
& \begin{tabular}[c]{@{}l@{}} Output \\feature\\maps\end{tabular} 
& \begin{tabular}[c]{@{}l@{}} Kernel\\Size \end{tabular}                  
& \begin{tabular}[c]{@{}l@{}} Input \\Width/Height\end{tabular}                                                     
& Pooling & \begin{tabular}[c]{@{}l@{}} Number\\of passes\end{tabular}                                                   
\\ \hline
1 & 3 & 16 & 1 & 224x224 & Yes & 1\\
2 & 16 & 16 & 7 & 112x112 & Yes & 1 \\
3 & 16 & 32 & 7 & 54x54 & Yes & 1\\
4 & 32 & 64 & 5 & 24x24 & No & 1 \\
5 & 64 & 64 & 5 & 22x22 & No & 1\\
6 & 64 & 64 & 5 & 20x20 & No & 1 \\
7 & 64 & 128 & 3 & 18x18 & No & 1\\
8 & 128 & 128 & 3 & 18x18 & No & 1 \\
9 & 128 & 128 & 3 & 18x18 & No & 1\\
10 & 128 & 128 & 3 & 18x18 & No & 1 \\
11 & 128 & 128 & 3 & 18x18 & Yes & 1\\
\hline
\end{tabular}
\end{table}

\subsection{VGG16 and VGG19 networks}
\label{sec:vgg19}

The VGG16 and VGG19 networks~\cite{Simonyan_Zisserman14} are large CNN
architectures used for classifying the ImageNet dataset. For 224x224x4 input
images, they require 31~GOp/frame and 39~GOp/frame respectively, and can be
trained to achieve 68.3\% and 71.3\% Top-1 accuracy using our ADaPTION tool.
For evaluating the hardware performance of NullHop reported in
Sec.~\ref{sec:results}, we used randomly chosen images from the dataset.
Because of the image size, these two networks require particular arrangements
so that they can be implemented on NullHop as discussed next:

\paragraph{Number of output feature maps} Since there are 128 MAC blocks and
at least one MAC block must be dedicated to each output feature map, multiple
passes through the input feature maps are needed in order to produce more
than 128 output feature maps. This is the case for example for layers 5-8 of
VGG19 which has 256 output feature maps. The output feature maps are divided
into multiple subsets with at most 128 maps per subset. Each subset is
produced in one pass through the input feature maps. The kernels for the
current subset of output feature maps are loaded into NullHop at the
beginning of each pass.

\paragraph{IDP memory size} If the input feature maps cannot fit into
NullHop's SRAM and there are more than 128 output feature maps, the input
feature maps must be streamed in multiple times. The probability of multiple
passes is lower in the zero-skipping mode since it is more likely that the
compressed input feature maps can fit into the accelerator SRAM. The 512 KB
of memory implemented in the IDP module was verified as sufficient to avoid
multiple streaming for all tested input images.

\paragraph{Kernel memory size per output feature map:} Each MAC block has
access to 8kB of kernel memory, i.e, 4k kernel values. If there are more than
4k kernel values associated with one output feature map as in layers 10-16 of
VGG19 (where each output feature map has $512\times 3\times 3=4608 $ kernel
values, which is larger than 4096), then multiple MAC blocks and their kernel
banks must be clustered together. Each cluster is responsible for producing one output feature map and must have
enough kernel memory to store the kernels for one output feature map. For
layers 10-16, the MAC blocks are clustered in groups of 2 to produce 64 output
feature maps for one pass through the input feature maps.

\subsection{Giga1Net}

Giga1Net is a 1~GOp/frame (about two-thirds the size of~\cite{Krizhevsky_etal12}) 
that we designed for stressing our accelerator. It contains various network layer configurations (Table~\ref{table:giganet}). Rather than achieving high accuracy as the main goal, the purpose of the Giga1Net architecture is to identify inefficiencies that arise in the hardware pipeline from the use of multiple combinations of different kernel sizes (1x1, 3x3, 5x5, 7x7), different number of output feature maps (from 16 to 128), the presence or absence of the pooling operation and zero padding.

\subsection{RoshamboNet}
\label{sec:roshambo}

RoshamboNet is a 5-layer, 20~MOp, 114k weight CNN architecture, described in
Table \ref{table:roshambonet}, trained to play the rock-scissors-paper
game~\cite{lungu_live_2017}. This network can classify input images of
size 64x64 obtained from the DVS of a DAVIS camera using the same training and
feature extraction stage approach from~\cite{moeys_steering_2016}. The
network outputs 4 classes: "rock", "scissors", "paper" or "background" from
each feature vector. Each frame is a 2D histogram of 2k DVS events.
Because of the asynchronous nature of the DVS output, the frame rate varies
from $<$1\,Hz to over 400\,Hz depending on the speed of the moving hand.

We connected NullHop to a DAVIS camera, a robot hand, and an LED display, driven by a
customized version of cAER running on the Zynq ARM processor. It was
demonstrated at NIPS 2016 Live Demonstration track, in software form at ISCAS~\cite{youtubeIulia,lungu_live_2017}, and four more public science events in 2017 and 2018. It beats human opponents by recognizing the player's symbol with over
99\% accuracy and reacting in less than 10ms to create a convincing illusion
of outguessing the opponent.

\subsection{Face Detector CNN}
\label{sec:face_detector_cnn}

The face detector is a small CNN designed to recognize whether a face is present or absent in an image obtained from the DAVIS camera. The DVS events are accumulated into 36x36 input images, again using the method of~\cite{moeys_steering_2016}. The network was trained on a dataset of 1800k frames collected from public face datasets and labeled DAVIS frames. The CNN architecture described in Table~\ref{table:face_detector} requires 1.98~MOp for classifying a single frame. 

\section{Results}
\label{sec:results}
The networks described in Sec.~\ref{sec:application_example} have all been run on both the Mentor QuestaSim HDL simulator and on our Xilinx Zynq platform. Results are summarized in Tables~\ref{table:rtl_results} and~\ref{table:fpga_results}.

\begin{table*}
\centering
\caption{RTL simulations results (convolutional layers only)}
\label{table:rtl_results}
\begin{tabular}{|l|c|c|c|c|c|c|c|c|}
\hline
\begin{tabular}[c]{@{}l@{}}Network\end{tabular}                                                                
& \begin{tabular}[c]{@{}l@{}}GOp/frame\end{tabular}                                                    
& \begin{tabular}[c]{@{}l@{}}ms/frame\end{tabular} 
& \begin{tabular}[c]{@{}l@{}}frame/s\end{tabular}
& \begin{tabular}[c]{@{}l@{}}GOp/s\end{tabular}                  
& \begin{tabular}[c]{@{}l@{}}Efficiency\end{tabular}                                                     
& \begin{tabular}[c]{@{}l@{}}GOp/s/W\end{tabular}                 
& \begin{tabular}[c]{@{}l@{}}MAC Utilization\end{tabular}                     
& \begin{tabular}[c]{@{}l@{}}MAC Utilization\\(no kernel loading time)\end{tabular}    
\\ \hline
VGG19  & 39.07 & 82.72 & 12.1 & 471.64 & 368.5\% & 3042.84 & 74.19\% & 97.87\% \\
VGG16 & 30.69 & 72.94 & 13.71 & 420.83 & 328.8\% & 2715.00 & 78.34\% & 98.14\% \\
Giga1Net & 1.040 & 4.16 & 240.07 & 249.68 & 195.1\% & 1610.79 & 67.31\% & 87.40\% \\
RoshamboNet & 0.018 & 0.2 & 4219.40 & 75.95 & 59.4\% & 490.00 & 33.58\% & 65.80\% \\
Face detector & 0.002 & 0.0264 & 37864.46 & 75.73 & 59.2\% & 488.57 & 40.90\% & 51.05\% \\ 
\hline
\end{tabular}      
\end{table*}

\begin{table*}
\centering
\caption{FPGA results (convolutions + fully connected + control overhead)}
\label{table:fpga_results}
\begin{tabular}{|l|c|c|c|c|c|c|}
\hline
\begin{tabular}[c]{@{}l@{}}Network\end{tabular}                                                                
& \begin{tabular}[c]{@{}l@{}}GOp/frame\end{tabular}                                                    
& \begin{tabular}[c]{@{}l@{}}ms/frame\end{tabular} 
& \begin{tabular}[c]{@{}l@{}}frame/s\end{tabular}
& \begin{tabular}[c]{@{}l@{}}GOp/s\end{tabular}                                                                  
& \begin{tabular}[c]{@{}l@{}}ms/frame\\(CNN Only)\end{tabular} 
& \begin{tabular}[c]{@{}l@{}}Efficiency\\(CNN Only)\end{tabular}     
\\ \hline
VGG19 & 39.017 & 2439 & 0.410 & 16.10 & 1819 & 143.40\% \\
VGG16 & 30.693 & 2269 & 0.441 & 17.196 & 1506 & 136.45\% \\
Giga1Net & 1.040 & 115.8 & 8.64 & 8.99 & 81.8 &  99.41\% \\
RoshamboNet & 0.018 & 5.49 & 182.15 & 3.28 & 5.98 & 22.23\% \\
Face detector &  0.002 & 3.289 & 304.05 & 0.61 & 0.57 & 23.31\%  \\ 
\hline
\end{tabular}
\end{table*}

\begin{table}
\centering
\caption{NullHop main features summary}
\label{table:nhp_summary}
\begin{tabular}{|l|c|}
\hline
    \textbf{Technology} &\begin{tabular}[c]{@{}l@{}}ASIC: GF 28nm\\FPGA: Xilinx Zynq 7100\end{tabular}\\
    \hline
    \textbf{Chip Size} & 8.1mm$^2$ \\
    \hline
    \textbf{Core Size} & 6.3mm$^2$ \\
    \hline

    \textbf{\#MAC} & 128 \\
    \hline
   \textbf{Clock Rate}&\begin{tabular}[c]{@{}l@{}}ASIC: 500 MHz\\FPGA: 60 MHz\end{tabular}\\
    \hline
    \textbf{Max. Effective Throughput}&\begin{tabular}[c]{@{}l@{}}ASIC: 471 GOp/s\\FPGA: 17.19 GOp/s\end{tabular}\\
    \hline
    \textbf{Max. Effective Efficiency}&\begin{tabular}[c]{@{}l@{}}ASIC: 368\%\\FPGA: 143\%\end{tabular}\\
    \hline
    \textbf{Max. GOp/s/W}&\begin{tabular}[c]{@{}l@{}}ASIC: 3042 GOp/s/W\\FPGA: 28.8 GOp/s/W\end{tabular}\\
      \hline
    \textbf{Arithmetic Precision} & \begin{tabular}[c]{@{}l@{}}16-bit fixed point multipliers\\32-bit fixed point adders\end{tabular} \\
\hline
\end{tabular}
\end{table}

\subsection{VGG19 and VGG16}
\label{subsect:resultvgg}
When big and deep networks with a high level of sparsity are computed on NullHop, it achieves more GOp/s than the ideal maximum (128 GOp/s, equal to the number of MAC units times their clock frequency) because operations are skipped, thereby achieving an efficiency greater than 100\%. VGG19 and VGG16, with respectively 471.64 GOp/s - 368.47\% efficiency and 420.83 GOp/s - 328.8 \% efficiency illustrate this effect of high sparsity. 

When processing each layer, there is an initial loading phase where the kernels are loaded into the the accelerator's kernel memory followed by the first $k$ input rows of the feature maps. After this initial loading phase, the controllers are activated to process the input pixels while the rest of the input feature maps are loaded in parallel. For all layers except the first one, the utilization of the 128 MAC blocks outside the initial loading phase was consistently above $99\%$, that is, in more than $99\%$ of clock cycles, each MAC block was carrying out a multiplication. The zero-skipping pipeline thus efficiently utilizes the compute resources even when faced with unpredictable sparsity patterns. This pipeline, though data-dependent, achieves a MAC utilization that is on par with dense processing pipelines that carry out all the MAC operations described in~\eqref{eq:conv} %Eq.~\ref{eq:conv} 
in a data-independent manner. 

The first convolutional layer in the VGG networks is special since the accelerator performance is limited by the bandwidth of the output bus from the accelerator to the external DRAM: each output pixel has a computational cost of $3\times 3\times 3 = 27$ MAC operations. The 128 MAC blocks can thus dispatch on average $128/27 = 4.7$ output pixels per cycle. However, the output bus can transmit at most $2$ non-zero pixels per cycle. Despite the sparsity of the output, the output bus still must slightly throttle the compute pipeline to be able to transmit the pixels. Thus, the MAC utilization of the first layer is $60\%$. 
 
RTL performances at 500 MHz show that the accelerator is able to process these 
networks at about 13\,FPS (frames/second), a value which is high enough for real time processing. The FPGA implementation, despite the lower frequency and the overhead due to using the ARM as the external controller, is able to process the VGG16 CNN in about 1.5s, a number that is aligned with state of the art.

\subsection{Giga1Net}

The first layer of Giga1Net is composed of sixteen 1x1x3 kernels, requiring 3 MAC operations for each output pixel. When processing 16 output feature maps, NullHop clusters 8 multipliers to work together on a single output map. Since each 1x1 kernel in the first layer requires only 3 MAC operations, 5 MAC units are idle during the computation. Furthermore, the write of the output pixels on the bus is again a bottleneck, since now the MACs are producing 16 output pixels per clock cycle. As result, the MAC utilization in the first layer is 10\%. For successive layers, the NullHop pipeline is able to adapt better to the architecture of the network, achieving about 250 GOp/s - 195\% efficiency. 

The FPGA implementation suffers from the low compute performance of the ARM CPU which computes the fully-connected layers. Despite that, the system is able to classify frames at more than 8~FPS running at only 60 MHz, with an efficiency for the full system of almost 100\%.

\subsection{RoshamboNet and Face Detector}
Both the RoshamboNet and Face Detector are 
small CNNs for which the Nullhop pipeline is 
pushed to its limits in terms of flexibility, 
representing the lower limit of NullHop efficiency. 
For such small networks, the main performance 
limitation lies in the I/O bandwidth: NullHop MACs 
are forced to be idle for about 50\% of the 
time while they wait for kernels or data to 
be loaded or previous results to be streamed out. 
Despite this bottleneck, NullHop achieves 
faster than real time performance at only 50\,MHz clock frequency.

\subsection{Memory Power Consumption Estimation}

To estimate the total power consumption of an ASIC implementation of 
our system including external DRAM memory access, 
we collected statistics about data movement 
during computation. We used a DRAM memory 
access energy of 21\,pJ/bit reported 
for an LPDDR3 memory model~\cite{Schaffner2015}, 
and ran each network as fast as possible. 
Table \ref{table:ddr_power} shows the results. For small networks, the number of operations 
necessary to compute the convolutions is low and 
the accelerator spends most of the time performing
memory transfer to load and store kernels and activations. 
The result is that both RoshamboNet and FaceNet 
have higher memory power consumption than the 
larger VGG16 and VGG19 networks. In the case of 
VGG16, the overall memory transfer is an 
average of 42\,MB/frame. The only available 
comparison is Eyeriss~\cite{Chen_etal16}, which 
reports 341\,MB for a batch of 3 VGG16 images. 
In batch mode, Eyeriss transfers about 113 MB/frame, 
which is almost 3 times more I/O than Nullhop.

\begin{table}[h]
\centering
\caption{Power estimation using LPDDR3}
\label{table:ddr_power}
\begin{tabular}{|l|c|c|c|}
\hline
\begin{tabular}[c]{@{}c@{}}Network\end{tabular}                                                                
& \begin{tabular}[c]{@{}c@{}}LPDDR3 Power\\ {[mW]}\end{tabular}                                                    
& \begin{tabular}[c]{@{}c@{}}Total Power\\ {[mW]}\end{tabular} 
& \begin{tabular}[c]{@{}c@{}}Power Efficiency\\ {[GOp/s/W]}\end{tabular}
\\ \hline
VGG19& 114 & 269 & 1751 \\
VGG16& 102 & 257 & 1634  \\
Giga1Net& 129 & 284 & 878 \\
RoshamboNet& 219 & 374 & 203  \\
FaceNet& 159 & 314 & 250   \\
\hline
\end{tabular}
\end{table}

\section{Discussion}

\subsection{Comparison with Prior Work}
\begin{table*}
\centering
\caption{Comparison with prior work}
\label{table:result_comparison}
\begin{tabular}{|l|c|c|c|c|c|c|c|c|}
\hline
\begin{tabular}[c]{@{}c@{}}Architecture\end{tabular}                                                                
& \begin{tabular}[c]{@{}c@{}}Core Size\\\relax  [mm\textsuperscript{2}]\end{tabular}                                                    
& \begin{tabular}[c]{@{}c@{}}Technology\end{tabular} 
& \begin{tabular}[c]{@{}c@{}}Maximum Frequency\\\relax[MHz]\end{tabular}
& \begin{tabular}[c]{@{}c@{}}Theoretical\\Peak Performance\\\relax[GOp/s]\textsuperscript{[1]}\end{tabular}   
& \begin{tabular}[c]{@{}c@{}}Effective Performance\\\relax[GOp/s]\textsuperscript{[2]}\end{tabular}  
& \begin{tabular}[c]{@{}c@{}}Efficiency\textsuperscript{[3]}\end{tabular}   
& \begin{tabular}[c]{@{}c@{}}Effective Power Efficiency\textsuperscript{[4]}\\\relax[GOp/s/W]\end{tabular}    
\\ \hline
NullHop\textsuperscript{[a]}& 6.3 & GF 28nm & 500  & 128 & 471 & 368\% & 1571 (3042\textsuperscript{[+]})\\
NullHop\textsuperscript{[a1]}& 6.3 & GF 28nm & 500  & 128 & 420 & 328\% & 1634 (2715\textsuperscript{[+]})\\
NullHop-FPGA\textsuperscript{[b]}& - & - & 60  & 15 & 16.1 & 106.95\% & 28.8 \\
NullHop-FPGA\textsuperscript{[b1]}& - & - & 60  & 15 & 17.2 & 114.24\% & 27.4 \\
\hline
Eyeriss\textsuperscript{[c]}& 12.25 & TSMC 65nm& 250  & 84 & 27  & 32\% & 115\textsuperscript{[+]}\\
Origami\textsuperscript{[d]}& 3.09 & UMC 65nm & 500  & 196 & 145  & 74\% & 437\textsuperscript{[+]}\\
%NeuFlow\textsuperscript{[e]}& 12.5 & IBM 45nm & 400  & 320 & 294 & 91\%  & 490\textsuperscript{[+]}\\
%Neuflow commented out because of ambiguity in the sources about performances
ShiDianNao\textsuperscript{[e]}& 4.86 & 65nm & 1000  & 128 & - & - & - \\
Moons et al.\textsuperscript{[f]}& 2.4 & 40nm LP & 204  & 102 & 71 & 69\% & 940\textsuperscript{[+]}\\
Envision\textsuperscript{[g]}& 1.87 & UTBB 28nm & 200  & 102 & 76 & 74\%  & 1000\textsuperscript{[+]}\\
Cambricon-X& 6.38 & TSMC 65nm & 1000  & 512 & 544 & 106\%  & 571\\

DNPU\textsuperscript{[l]} & 16 & 65nm & 200 & 300  & 270 & 90\% & 4200\textsuperscript{[m,+]}\\
Yin et al.\textsuperscript{[n]} & 19.36 & TSMC 65nm & 200  & 409.6 & 368.4 & 90\% & 1027\\
UNPU\textsuperscript{[o]} & 16 & 65nm & 200  & -\textsuperscript{[p]} & 345.6 & - & 3080\textsuperscript{[+]}\\

\hline
\end{tabular}
\begin{tablenotes}
\setlength{\columnsep}{0.8cm}
\setlength{\multicolsep}{0cm}
\begin{multicols}{3}
      \tiny
       \item \textsuperscript{[1]}Computed as number of MAC/SOP units times clock frequency
       \item \textsuperscript{[2]}Measured performance
       \item \textsuperscript{[3]}Effective GOp/s divided by theoretical peak performance
       \item \textsuperscript{[4]}Including memory power consumption. Marker \textsuperscript{[+]} indicates core-only power consumption or not specified.

            \item\textsuperscript{[a]}VGG19 - Convolutional layers only
            \item\textsuperscript{[a1]}VGG16 - Convolutional layers only
              \item\textsuperscript{[b]}VGG19 - Full system
              \item\textsuperscript{[b1]}VGG16 - Full system
            \item\textsuperscript{[c]}VGG16 with batch size = 3 - Full system\cite{Chen_etal16} 
             \item\textsuperscript{[d]}Custom CNN - Full system\cite{Cavigelli_etal15} 
             \item \textsuperscript{[e]}Multiple Custom CNN - Convolutional layers only\cite{Du2015}  
            \item \textsuperscript{[f]}AlexNet - Convolutional layers only\cite{Moons2016} 
            \item \textsuperscript{[g]}VGG16 - Convolutional layers only\cite{Moons2017} 
              \item \textsuperscript{[l]}Value computed using the reported fps for 16-bit AlexNet\cite{Shin2017}
               \item \textsuperscript{[m]}Device working at low frequency (100 Mhz)\cite{Shin2017}
                \item \textsuperscript{[n]}AlexNet - Full system\cite{Yin2017}
               
                 \item \textsuperscript{[o]}Unspecified 16-bit precision network\cite{Lee18}
                  \item \textsuperscript{[p]}Serial multiplier implementation\cite{Lee18}
\end{multicols}
\end{tablenotes}
\end{table*}

Table~\ref{table:result_comparison} provides a summary of NullHop results and
comparison to prior CNN accelerators, showing how in ASIC simulations the
system is able to achieve state-of-the-art performance. The most relevant
result obtained is in terms of efficiency. While other architectures
suffer significant discrepancies between the theoretical peak performances and
the effective ones, with exception of~\cite{Zhang2016}, NullHop is able
to achieve an efficiency consistently higher than 100\% (and up to 368\% for a
large CNN), because of its zero-skipping pipeline and high MAC utilization.
It confirms the validity of the concept of sparse computation
introduced for fully-connected layers in~\cite{Han2016} and, for the
first time, takes it to the convolutional layers that constitute the bulk
of CNN computation. This advantage does not have any drawback
since all other main specifications -- core size, frequency and number of MAC
units -- are similar or better than other proposed solutions. The only ASIC with higher throughput is Cambricon-X~\cite{Zhang2016}, which achieves a remarkable 544 GOp/s in only 6.3\,mm$^2$, but implements twice the number of multipliers and runs at twice the
frequency of NullHop. However, Cambricon-X is about 3 times less power efficient than NullHop.

Both implementations in~\cite{Shin2017} and~\cite{Lee18} have a higher power efficiency than Nullhop, but provide consistently lower performances (\textless 350 GOp/s) using more MAC units. They also require a larger area (16\,mm$^2$), but this is justified by their support for Recurrent Neural Networks and variable bit precision.

\subsection{Conclusion}
Sec.~\ref{subsect:resultvgg} showed 
that the utilization of MACs in NullHop is consistently above $99\%$
if not limited by the input or output bandwidth.
NullHop allows sparse computation with high utilization
comparable to pipelines operating on dense representations.
Thus, it achieves a speedup directly proportional
to a CNN's sparsity since it does not waste cycles on zero input pixels,
while preserving a high level of CNN architecture flexibility. NullHop has quite high precision of 16-bit
weights and states, making it easy to train
CNNs with acceptable accuracy 
compared with the difficult CNN hyperparameter
tuning and extended training time required by 
super-reduced precision networks, e.g.~\cite{mishra_wrpn:_2017}. Even so, a dedicated IC implementation of NullHop would achieve state-of-art power efficiency of 3~TOp/s/W.

Sec.~\ref{sec:compression} showed that the NullHop sparsity map compression achieves a
higher compression ratio than run-length encoding schemes
that were previously used to compress input feature maps and
allows direct operation} on their compressed representation. This compression reduces I/O power consumption that is a crucial component of system power consumption and computing time, making NullHop suited for embedded systems applications.

NullHop sprang from the fundamental neural organizing principle of sparse computation resulting from threshold-linear neural activation~\cite{hahnloser_digital_2000, Nair_Hinton10,glorot_deep_2011}. Only later did it become clear that exploiting sparsity also
allows large reductions in power consumption, memory access, and compute time, all of which are key to deployment for real world applications. 
The NullHop data encoding and processing pipeline are 
optimized to handle spatially-sparse data representations 
with a flexible range of input and output feature maps. 
NullHop was developed specifically for the case of CNNs. 
However, its compression scheme can also be 
used to encode sparse data in other types of networks. 
Exploiting \textit{temporal} sparsity
can greatly reduce recurrent neural network memory access
~\cite{neil_delta_2017,gao_deltarnn:_2018}. 
Future architectures will certainly achieve even higher efficiency by 
combining these spatial and temporal sparsity principles 
with compression and controlled precision, even without departing from
the immense practical advantages of synchronous logic.

\section*{Acknowledgments}
This work was supported by Samsung Advanced Inst. of Technology (SAIT), the University of Zurich and ETH Zurich.

\bibliographystyle{IEEEtran}
\bibliography{biblio,tobi,ricardo,moritz,ale}

%\vskip -3\baselineskip plus -1fil
\vspace{-1.5in}
\vspace{.1in}

\begin{IEEEbiography}
    [{\includegraphics[width=1in,height=1.25in,clip,keepaspectratio]{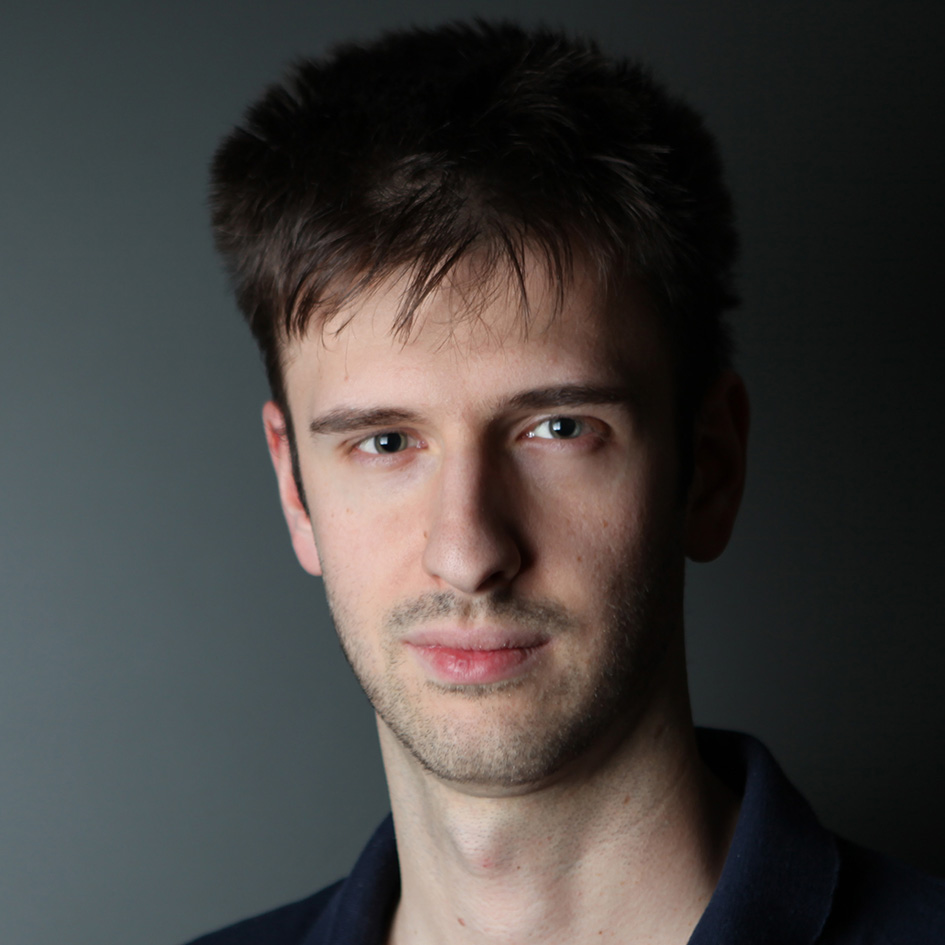}}]{Alessandro Aimar}
received the B.Sc. degree in Physical Engineering from Politecnico di Torino (Italy) and the M.Sc. degree in Nanotechnologies from a joint program of Politecnico di Torino, INP Grenoble (France) and EPFL (Switzerland). After working as engineer at Imagination Technologies (UK) he joined the Institute of Neuroinformatics for his PhD. In 2017 he founded Synthara Technologies, a startup focused on deep learning and neuromorphic hardware.
\end{IEEEbiography}

\vskip -3.75\baselineskip plus -1fil

\begin{IEEEbiography}
    [{\includegraphics[width=1in,height=1.25in,clip,keepaspectratio]{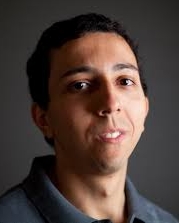}}]{Hesham Mostafa}
obtained his masters in electrical engineering from the Technical University of Munich in 2010, and his PhD in Neuroinformatics from the Institute of Neuroinformatics in 2016. He is currently a post-doc at the integrated systems neuro-engineering lab at the Institute of Neural Computation at UC San Diego.\end{IEEEbiography}

\vskip -3\baselineskip plus -1fil

\begin{IEEEbiography}
[{\includegraphics[width=1in,height=1.25in,clip,keepaspectratio]{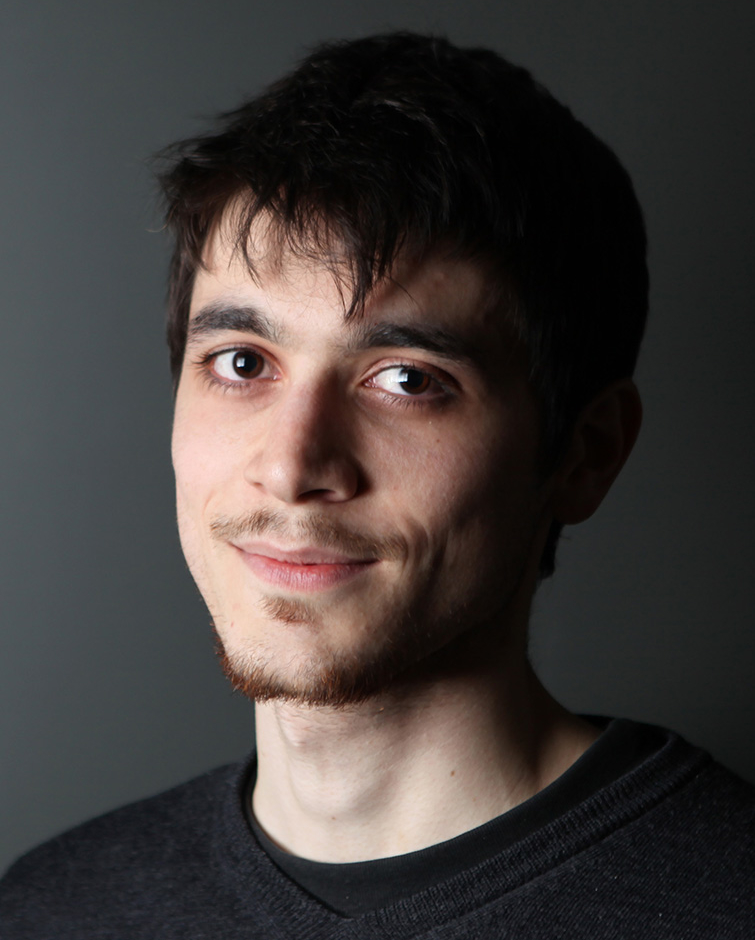}}]{Enrico Calabrese}
obtained the B.Sc. degree in physics from Universita' Degli Studi di Ferrara, and the joint M.Sc. degree in micro and nanotechnologies for integrated systems from Politecnico di Torino, Grenoble INP and EPFL. He is currently a PhD candidate at the Institute of Neuroinformatics.
His research interests include Deep Learning algorithms for artificial vision and attention.\end{IEEEbiography}

\vskip -3\baselineskip plus -1fil

\begin{IEEEbiography}
    [{\includegraphics[width=1in,height=1.25in,clip,keepaspectratio]{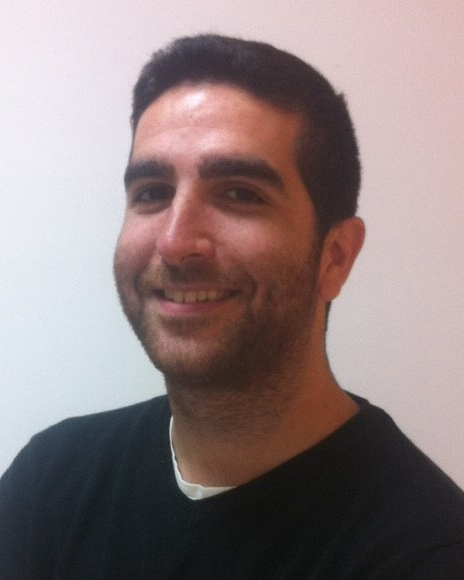}}]{Antonio Rios-Navarro}
received the B.S. degree in computer science engineering, the M.S. degree in computer engineering, and the PhD degree in neuromorphic engineering from the University of Seville, Seville, Spain, in 2010, 2011, and 2017, respectively. He is currently a post-doc at the Computer Architecture and Technology department, University of Seville. His current research interests include neuromorphic systems, real-time spikes signal processing, FPGA design and Deep Learning.
\end{IEEEbiography}

\vskip -3.65\baselineskip plus -1fil

\begin{IEEEbiography}
    [{\includegraphics[width=1in,height=1.25in,clip,keepaspectratio]{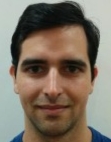}}]{Ricardo Tapiador-Morales}
received the B.S. degree in Computer Sciences in 2015 and the M.S. degree in Computer Engineering in 2016, from the University of Seville, Spain.He is currently a PhD student of Computer Architecture and Technology department,University of Seville. His research interests include neuromorphic engineering, convolutional neural networks, FPGA design and embedded systems.
\end{IEEEbiography}

\vskip -3\baselineskip plus -1fil

\begin{IEEEbiography}
    [{\includegraphics[width=1in,height=1.25in,clip,keepaspectratio]{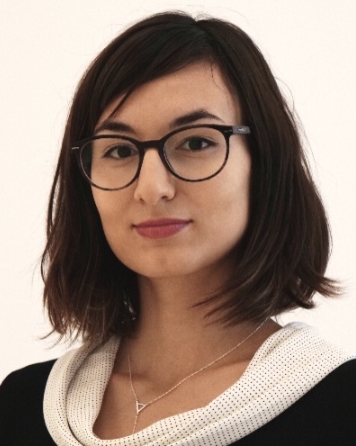}}]{Iulia-Alexandra Lungu}
received her BSc in Bioinformatics from the University Claude Bernard Lyon and her MSc in Computational Neuroscience from the Technical University Berlin. She is now pursuing her PhD degree at the Inst. of Neuroinformatics, where she works with talented hardware designers to develop efficient and powerful algorithms and hardware solutions for AI. Her main interests revolve around incremental and reinforcement learning. 
\end{IEEEbiography}

\vskip -3\baselineskip plus -1fil

\begin{IEEEbiography}
    [{\includegraphics[width=1in,height=1.25in,clip,keepaspectratio]{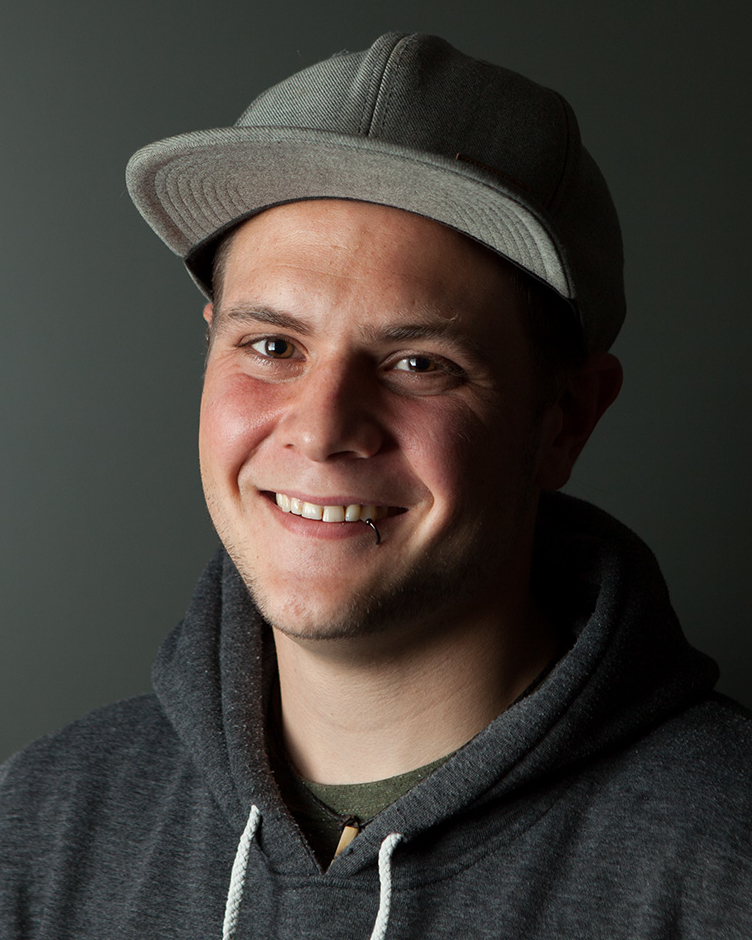}}]{Moritz B. Milde}
received the B.Sc. degree in biomimetics from Westphalian University of Applied Sciences and the M.Sc. in neurobiology from Bielefeld University. He is currently a PhD candidate at the Institute of Neuroinformatics, where he focuses on event-driven vision-based scene understanding for robotic navigation.
\end{IEEEbiography}

\vskip -3.1\baselineskip plus -1fil

\begin{IEEEbiography}
    [{\includegraphics[width=1in,height=1.25in,clip,keepaspectratio]{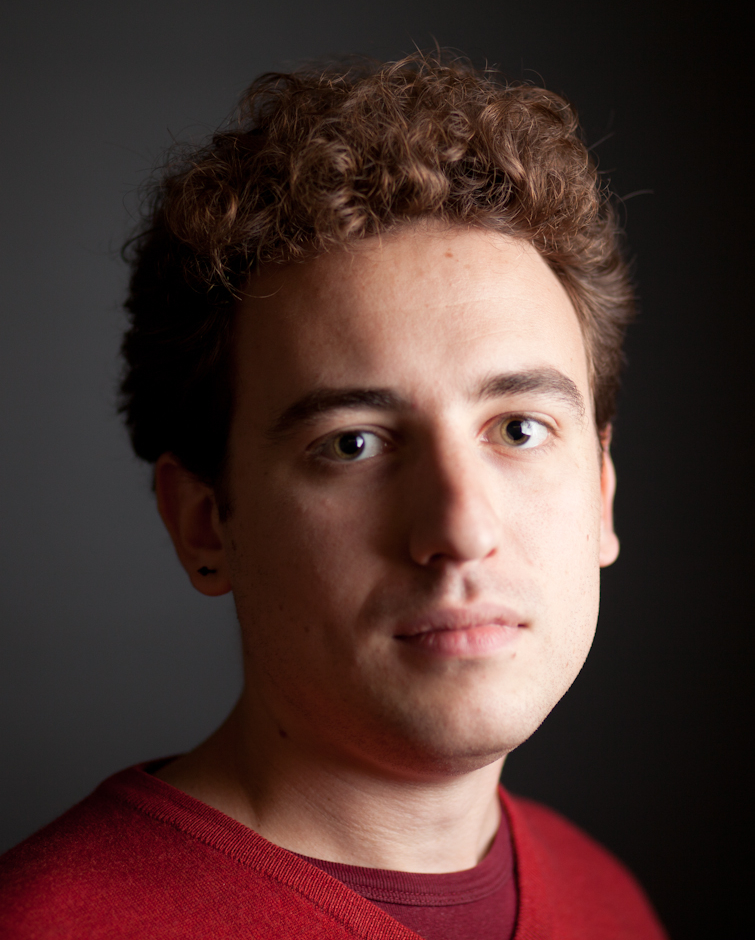}}]{Federico Corradi}
(M'13) received the B.Sc. degree in physics from Universita Degli Studi di Parma, Italy, the M.Sc. degree (cum laude) in physics from La Sapienza University, Rome, Italy, and the Ph.D. degrees in Natural Sciences from the University of Zurich, and in neuroscience from Neuroscience Center Zurich, Switzerland. He is currently with imec in Holland.
\end{IEEEbiography}

\vskip -3\baselineskip plus -1fil

\begin{IEEEbiography}
    [{\includegraphics[width=1in,height=1.25in,clip,keepaspectratio]{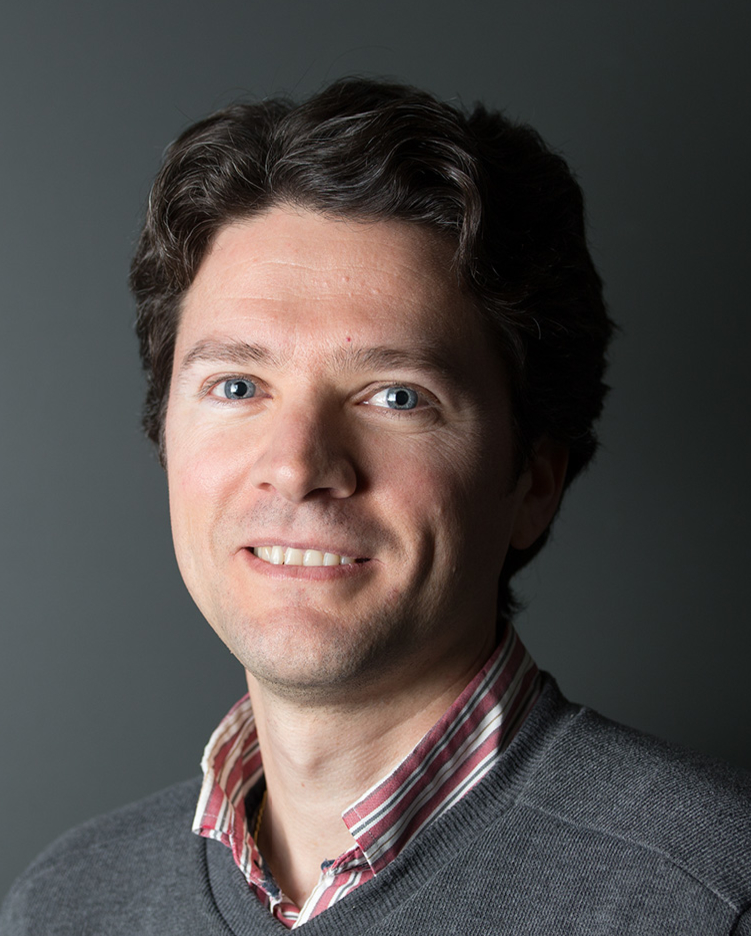}}]{Alejandro Linares-Barranco} 
(M'04-SM'17) received the B.S. degree in computer engineering, the M.S. degree in industrial computer engineering, and the Ph.D. degree in computer engineering (specialized in computer interfaces for neuromorphic systems) from the University of Seville in 1998, 2002, and 2003, respectively. He is Associate Professor in the University of Seville since 2009 and head of the Architecture and Computer Tech. Dept. 
\end{IEEEbiography}

\vskip -3.1\baselineskip plus -1fil

\begin{IEEEbiography}
    [{\includegraphics[width=1in,height=1.25in,clip,keepaspectratio]{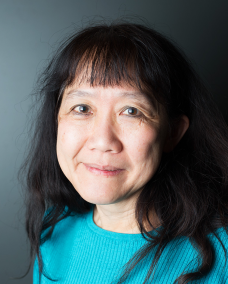}}]{Shih-Chii Liu} (M'02–SM'07) studied electrical engineering as an undergraduate at the Massachusetts Institute of Technology and received the Ph.D. degree in the computation and neural systems program from Caltech in 1997. She worked at various companies including Gould American Microsystems, LSI Logic, and Rockwell International Research Labs. Currently, she is a group leader at the Institute of Neuroinformatics. 
\end{IEEEbiography}

\vskip -3\baselineskip plus -1fil

\begin{IEEEbiography}
    [{\includegraphics[width=1in,height=1.25in,clip,keepaspectratio]{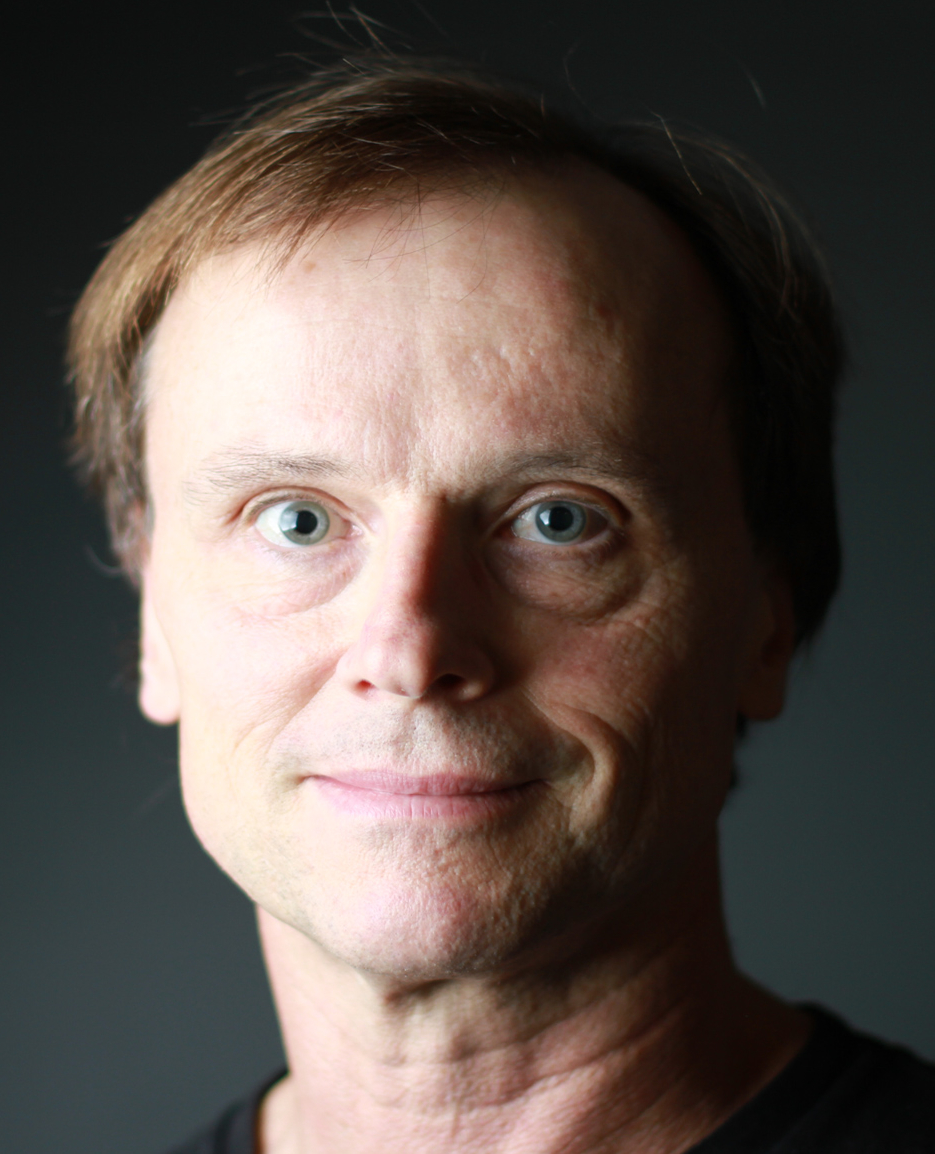}}]{Tobi Delbruck}
(M'99–SM'06–F'13) received the B.Sc. degree in physics from UC San Diego in 1986 and PhD degree from Caltech in 1993. He is currently a Professor of Physics and Electrical Engineering at ETH Zurich with the Institute of Neuroinformatics, where he has been since 1998. His group with SC Liu focuses on neuromorphic sensory processing and efficient deep learning.
\end{IEEEbiography}

\vspace{-1.5in}

\end{document}